\documentclass[lettersize,journal]{IEEEtran}
\usepackage{amsmath,amsfonts}
\usepackage{algorithmic}
\usepackage{algorithm}
\usepackage{array}
\usepackage[caption=false,font=normalsize,labelfont=sf,textfont=sf]{subfig}
\usepackage{textcomp}
\usepackage{stfloats}
\usepackage{url}
\usepackage{verbatim}
\usepackage{graphicx}
\usepackage{cite}

\usepackage{booktabs}
\usepackage{makecell} 
\usepackage{xcolor}
\usepackage{rotating}

\begin{document}

\title{From Pixels to Titles: Video Game Identification by Screenshots using Convolutional Neural Networks}

\author{Fabricio Breve
\thanks{F. Breve is with São Paulo State University (UNESP), Institute of Geosciences and Exact Sciences, Rio Claro, São Paulo, Brazil}}


\maketitle

\begin{abstract} 
This paper investigates video game identification through single screenshots, utilizing ten convolutional neural network (CNN) architectures (VGG16, ResNet50, ResNet152, MobileNet, DenseNet169, DenseNet201, EfficientNetB0, EfficientNetB2, EfficientNetB3, and EfficientNetV2S) and three transformers architectures (ViT-B16, ViT-L32, and SwinT) across 22 home console systems, spanning from Atari 2600 to PlayStation 5, totalling 8,796 games and 170,881 screenshots. Except for VGG16, all CNNs outperformed the transformers in this task. Using ImageNet pre-trained weights as initial weights, EfficientNetV2S achieves the highest average accuracy (77.44\%) and the highest accuracy in 16 of the 22 systems. DenseNet201 is the best in four systems and EfficientNetB3 is the best in the remaining two systems. Employing alternative initial weights fine-tuned in an arcade screenshots dataset boosts accuracy for EfficientNet architectures, with the EfficientNetV2S reaching a peak accuracy of 77.63\% and demonstrating reduced convergence epochs from 26.9 to 24.5 on average. Overall, the combination of optimal architecture and weights attains 78.79\% accuracy, primarily led by EfficientNetV2S in 15 systems. These findings underscore the efficacy of CNNs in video game identification through screenshots. 
\end{abstract}


\section{Introduction}

\IEEEPARstart{H}{umans} possess the remarkable ability to easily recognize their favorite video games or titles they have played frequently from a single screenshot. This proficiency is rooted in the presence of consistent visual elements, including sprites, heads-up displays (HUDs), and distinctive game scenarios. However, extending this capability to the automated identification of random video games from an extensive console library presents a challenge, even for the most dedicated gamers. Therefore, the concept of automatically identifying video games from single screenshots holds significant interest, not only for its technical complexity but also for its vast practical applications.

Automated game identification could offer significant benefits to various sectors within the gaming industry. Video game databases, search engines, and online platforms could gain considerably from this technology. By analyzing user-uploaded screenshots, these platforms can automatically generate metadata, including game titles, release dates, and developer information. Such automation would not only improve the accuracy of their game libraries but also enhance cataloging efficiency. Moreover, online streaming platforms could leverage screenshot recognition to provide real-time information to viewers about the games being played during live streams, enhancing the overall viewer experience. This technology opens doors to further innovation within the gaming ecosystem, potentially influencing game recommendation systems and aiding game-related research.

Most prior research on video game classification has centered around identifying their genre \cite{Souza2016,Goring2020,Suatap2020,Jiang2023}. However, this paper takes a more specific approach, focusing on classifying video game titles based on single screenshots using CNN and transformer models. The hypothesis is that their inherent capacity of automatically extracting relevant features from images is sufficient to identify video game titles from single screenshots in most scenarios, without relying on other features. To begin this research, a dataset encompassing 170,881 screenshots from 8,796 games of 22 popular home console systems was curated. The screenshots were sourced from the Moby Games Database \cite{MobyGames2023}. The proposed dataset spans a wide spectrum of gaming history, ranging from iconic consoles like the `Atari 2600' of the second generation to the cutting-edge `PlayStation 5' and `Xbox Series X/S' of the current generation, carefully selecting the most sold consoles from each generation between them. 

To tackle this task, well-established CNN architectures were selected: VGG \cite{Simonyan2015}, ResNet \cite{He2016}, MobileNet \cite{Howard2017}, DenseNet \cite{Huang2017}, and EfficientNet \cite{Tan2019,Tan2021}. These architectures have consistently demonstrated outstanding performance in previous works with different kinds of images \cite{Breve2022,Breve2023}, making them prime candidates for this game title classification task. Transformers are also being successfully employed in image classification; therefore, two of them were selected for this task: Vision Transformer \cite{Dosovitskiy2021} and Swin Transformer \cite{Liu2021}. The initial weights of all those models were first initialized with pre-trained weights from the ImageNet dataset \cite{ILSVRC15}, a widely adopted approach in transfer learning \cite{Oquab2014}. Subsequently, the fine-tuned weights from another dataset of screenshots were employed to enhance both classification accuracy and reduce training times of the best methods for this task. To the best of our knowledge, this marks the first attempt to tackle the challenge of game title classification using CNNs and transformers. It is worth noting that after training a network to identify game titles, incorporating new titles and screenshots into the dataset — whether they are added to the Moby Games Database or provided by the publisher — would necessitate adding additional nodes to the output layer and performing at least a few more training epochs to adjust the new network weights and fine-tune the convolutional layers. By pushing the boundaries of automated video game identification, the aim is to contribute valuable insights to game-related research and practical applications in the ever-evolving gaming industry.

The remainder of this paper is organized as follows. Section~\ref{sec:RelatedWork} presents related work on video game classification. Section~\ref{sec:Dataset} presents the Moby Games Database and how the dataset was sourced from it. Section~\ref{sec:CNNArchitectures} shows the CNN and transformer architectures employed in this paper. Section~\ref{sec:CNNComparison} displays the experiments comparing the CNN and transformer architectures in the task of identifying the games from their screenshots, initialized with pre-trained weights from the ImageNet dataset. Section~\ref{sec:AlternativeWeights} demonstrates experiments using weights fine-tuned in another screenshots dataset, comparing the accuracy and training epochs with those obtained with the ImageNet weights and random weights. Section~\ref{sec:Limitations} presents some limitations of our proposed approach. Finally, the conclusions are drawn in Section~\ref{sec:Conclusions}.

\section{Related Work}
\label{sec:RelatedWork}

Most of the video games classification attempts so far aimed at genre classification. Souza et al. \cite{Souza2016} pioneering work classified game genre of gameplay videos. Their dataset comprises 700 gameplay videos spanning seven distinct game genres. In their research, they introduced novel descriptors known as Bossa Nova and BinBoost. The experimental outcomes demonstrated the effectiveness of their proposed approach, achieving an accuracy rate of 89.84\%. 

Göring et al. \cite{Goring2020} also introduced a novel method for classifying video games genres based on content. They used a dataset comprising 351 gameplay videos spanning six different genres. They employed random forest and gradient boosting trees as underlying machine-learning techniques, combined with feature selection of image-based features and motion-based features. The most promising results were achieved using the random forest classifier, which yielded an accuracy rate of 60.6\%.

Zadtootaghaj et al. \cite{Zadtootaghaj2018} introduced a game classification method based on graphical and video complexity. Their approach categorizes games into three distinct classes: low-complexity, medium-complexity, and high-complexity games. To achieve this classification, they developed a decision tree capable of accurately assigning a game to its appropriate complexity class with an accuracy rate of 96\%. The classification process relies on the analysis of specific attributes within the gameplay video, including the observation of a static area, assessment of the degree of freedom (DoF), and quantification of camera movement.

While the majority of classification endeavors have typically focused on broader categories, there was a unique attempt to classify video games by their titles more than a decade ago. In this pioneering effort, Madani et al. \cite{Madani2012} explored several fusion methods using a dataset containing 120,000 gameplay videos, with the objective of identifying game titles. Their approach integrated both audio and visual features to accurately pinpoint these specific game titles, ultimately achieving a F1-score of 0.82. Although their dataset is considerably large, they explored only a small number of games, identifying 30 distinct game titles. To the best of our knowledge, there has not been any other attempt to identify video games by their titles, especially in larger datasets with hundreds or thousands of titles, neither using video nor screenshots.

CNNs have significantly influenced the landscape of automatic image classification, including game classification. Suatap and Patanukhom \cite{Suatap2020} used games screenshots and icons provided in game stores to classify them by genre using convolutional neural networks and ensemble techniques. They achieved 40.3\% and 46.7\% classification accuracies for single icon and screenshot classification tasks, respectively. They increased these results to 40.5\% and 47.6\%, respectively, in a later work \cite{Suatap2022}, in which they also used features extracted from their trained models to perform other two tasks: similar game searching and quality assessment of game images based on the correctness of viewers' understanding of game content. 

Recently, Jiang and Zheng \cite{Jiang2023} devised deep neural networks for the purpose of classifying game genres using either cover images or description text. Their dataset encompassed cover images and description texts sourced from a pool of 50,000 games, which they categorized into 15 distinct genres. In their approach, several pre-trained CNNs were fine-tuned for the cover image classification task. For the classification of description text, they employed Long Short-Term Memory (LSTM) networks and the Universal Sentence Encoder (USE). The image-based model yielded a highest accuracy rate of 31.4\% when utilizing ResNet-50. They also achieved significant improvement in accuracy, up to 49.9\%, by combining image and text features within a multi-modal model. Both of these previous studies were limited to identifying game genres, and their accuracy was only moderate (below 50\%). Therefore, identifying game titles, which are more specific than genres, is still an open challenge. 

Video streaming platforms such as \emph{YouTube} and \emph{Twitch} have the capability to automatically identify games being played during live streams. However, this functionality appears to be currently limited to a few specific games. In most cases, streamers or uploaders still need to manually label the games they are playing. Unfortunately, these platforms do not openly disclose the methods they use for game identification, whether from existing literature or proprietary development.

\section{The Dataset}
\label{sec:Dataset}

The Moby Games Database \cite{MobyGames2023}, as stated on their website, is a project with the primary goal of cataloging comprehensive information about electronic games, encompassing computer, console, and arcade titles, on a game-by-game basis. This extensive catalog includes release details, credits, cover art, player-uploaded screenshots with captions, neutral descriptions, and much more. The database boasts a collection of over one million screenshots, organized by game titles and systems. Additionally, they offer an API that simplifies the process of requesting and retrieving dataset entries and screenshot files. Given these advantages, the Moby Games Database was selected as the source for the screenshots used in this research.

To maintain a focused scope for this initial endeavor in video game identification using CNNs, the study exclusively considered home video game consoles. Handheld devices, arcade games, and computer-based titles will be addressed in future research. The selection process involved choosing the top 22 best-selling home video game consoles of all time \cite{Wikipedia2023}. These consoles originate from six different manufacturers and exhibit varying quantities of screenshots per game in the database, ranging from none to a few dozen. In this paper, all experiments employed $k$-fold cross-validation with $k=5$. To ensure that each game has at least one screenshot in each of the five folds, only games with a minimum of five available screenshots were selected. Additionally, during the training phase, it was also assured that each game had screenshots in both the training and validation subsets.

\begin{figure*}
  \centering
    \subfloat[Enduro - Atari 2600]{\includegraphics[width=0.19\textwidth]{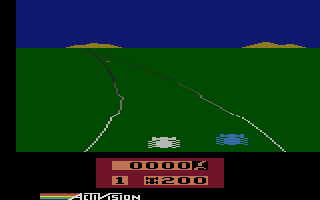}} \hfill
    \subfloat[Super Mario Bros. - NES]{\includegraphics[width=0.19\textwidth]{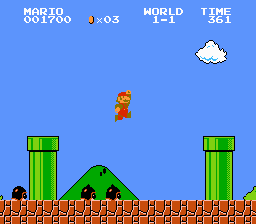}} \hfill
    \subfloat[The Legend of Zelda - NES]{\includegraphics[width=0.19\textwidth]{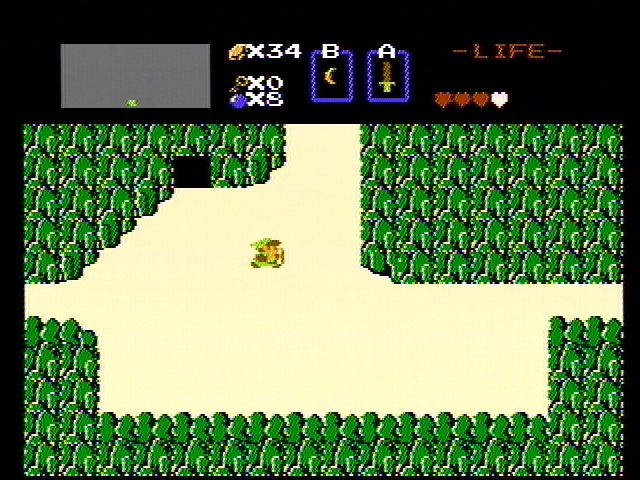}} \hfill
    \subfloat[Alex Kidd in Miracle World - Master System]{\includegraphics[width=0.19\textwidth]{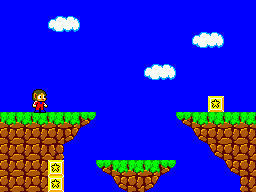}} \hfill
    \subfloat[Bonk's Adventure - PC Engine]{\includegraphics[width=0.19\textwidth]{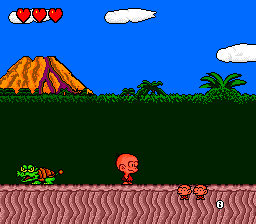}} 
    
    \subfloat[Sonic the Hedgehog - Genesis]{\includegraphics[width=0.19\textwidth]{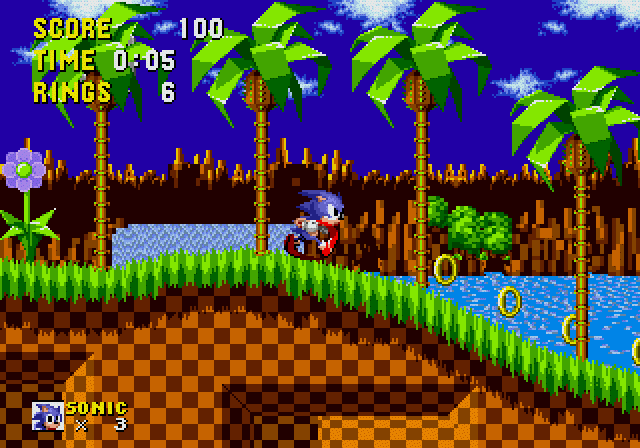}} \hfill
    \subfloat[Super Mario World - Super Nintendo]{\includegraphics[width=0.19\textwidth]{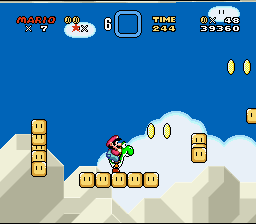}} \hfill
    \subfloat[Virtua Fighter 2 - Sega Saturn]{\includegraphics[width=0.19\textwidth]{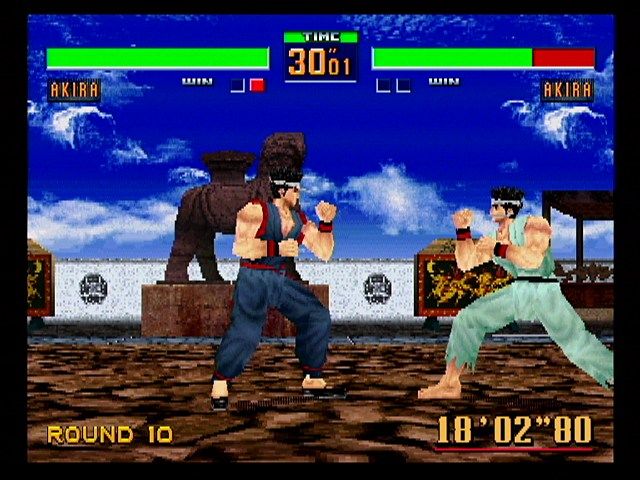}} \hfill
    \subfloat[Gran Turismo - PlayStation]{\includegraphics[width=0.19\textwidth]{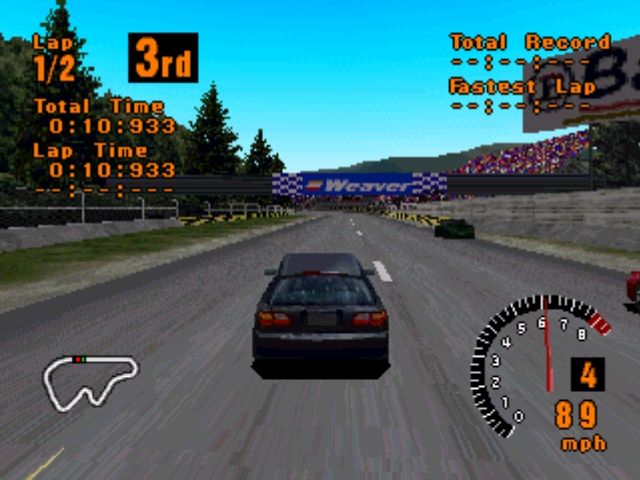}} \hfill
    \subfloat[Final Fantasy VII - PlayStation]{\includegraphics[width=0.19\textwidth]{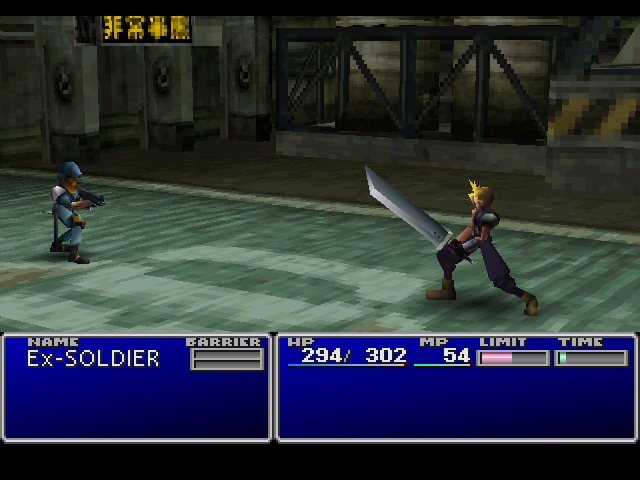}} 
    
    \subfloat[Super Mario 64 - Nintendo 64]{\includegraphics[width=0.19\textwidth]{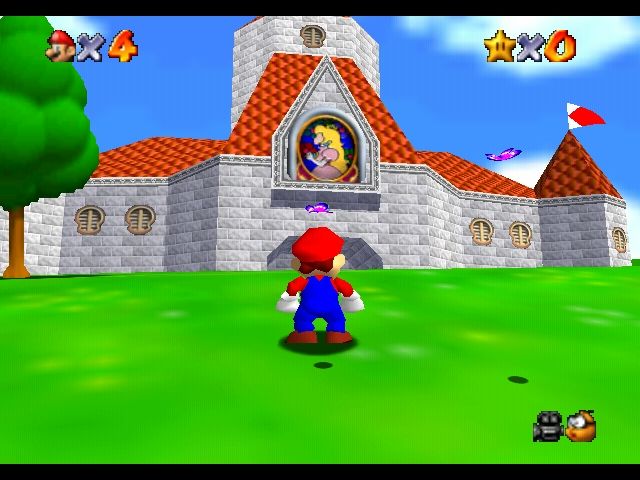}} \hfill
    \subfloat[Sonic Adventure - Dreamcast]{\includegraphics[width=0.19\textwidth]{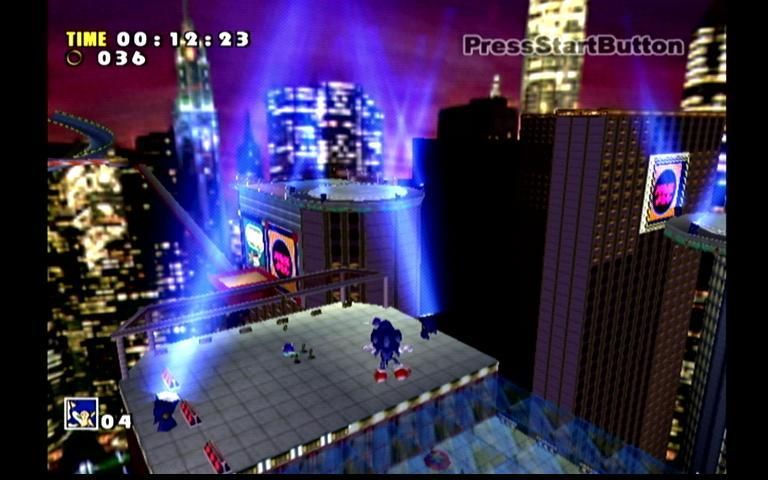}} \hfill
    \subfloat[Grand Theft Auto: San Andreas - PlayStation 2]{\includegraphics[width=0.19\textwidth]{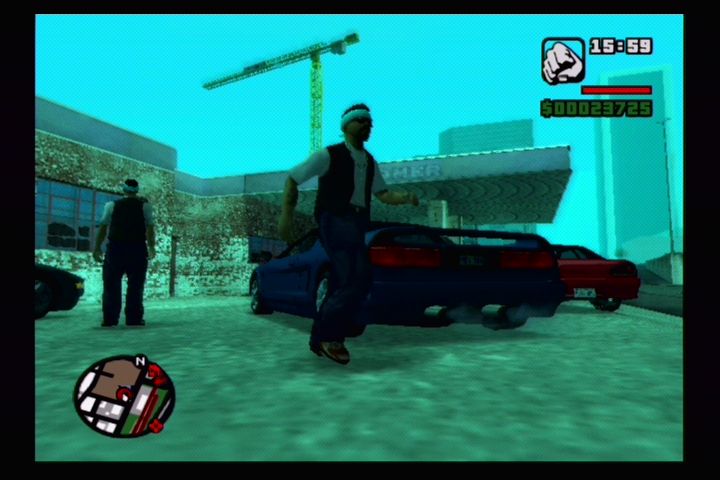}} \hfill
    \subfloat[Super Smash Bros.: Melee - GameCube]{\includegraphics[width=0.19\textwidth]{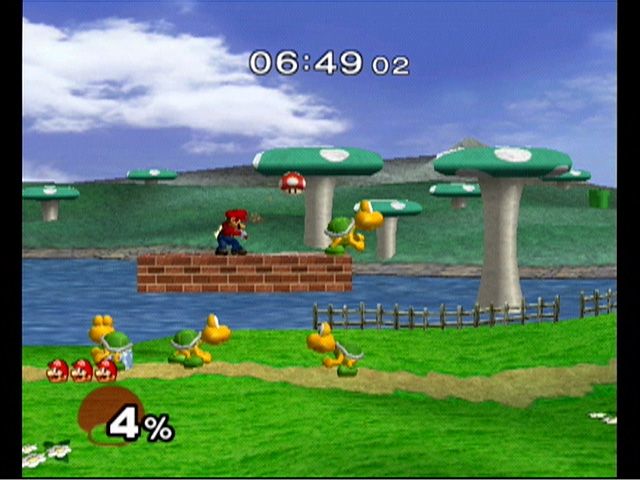}} \hfill
    \subfloat[Halo: Combat Evolved - Xbox]{\includegraphics[width=0.19\textwidth]{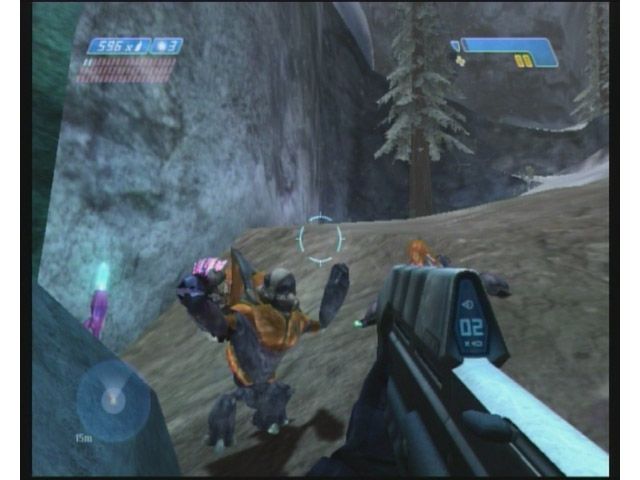}} 
    
    \subfloat[Halo 3 - Xbox 360]{\includegraphics[width=0.19\textwidth]{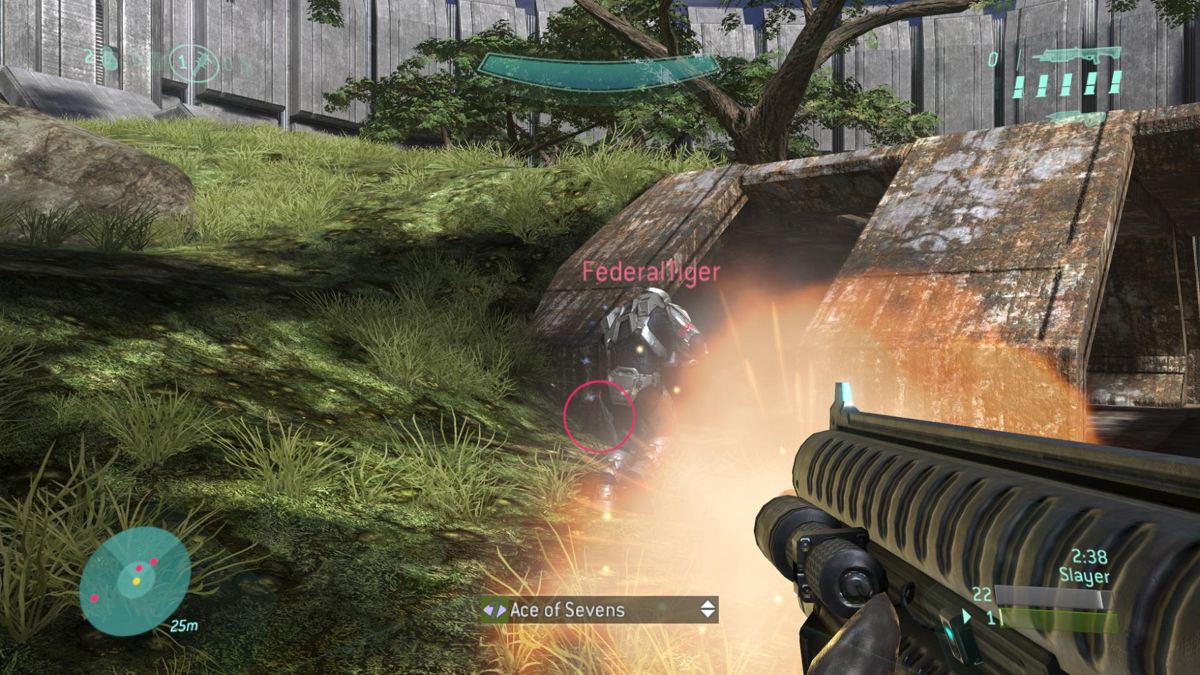}} \hfill
    \subfloat[The Last of Us - PlayStation 3]{\includegraphics[width=0.19\textwidth]{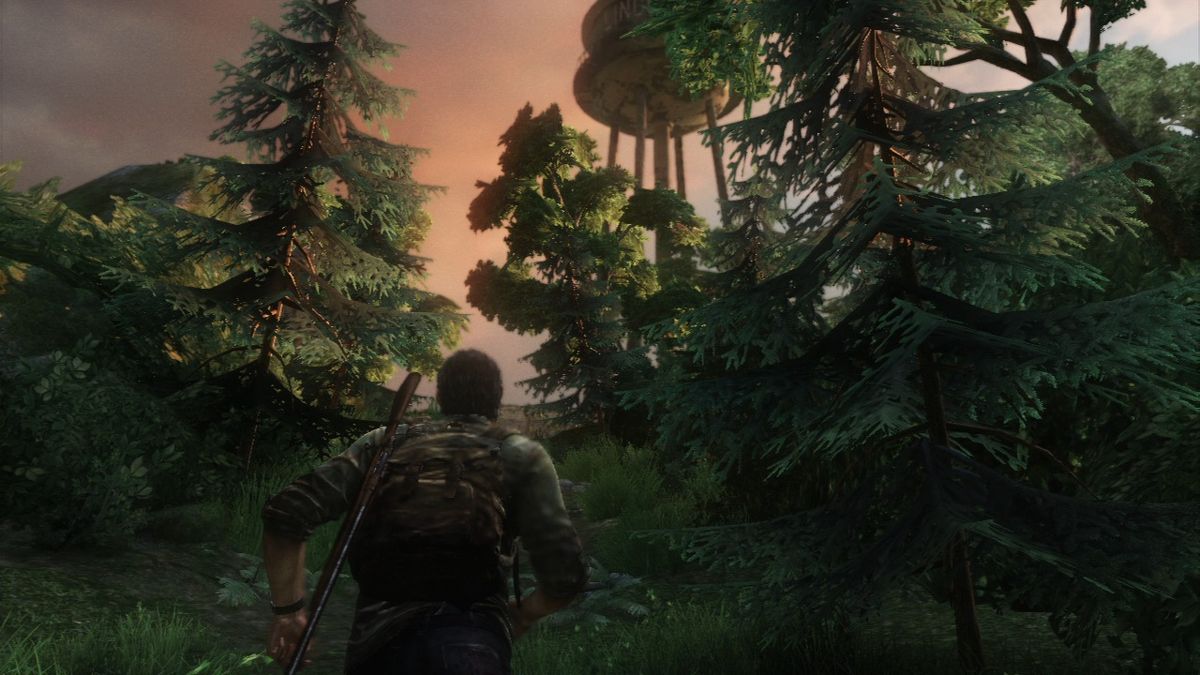}} \hfill
    \subfloat[Wii Sports - Wii]{\includegraphics[width=0.19\textwidth]{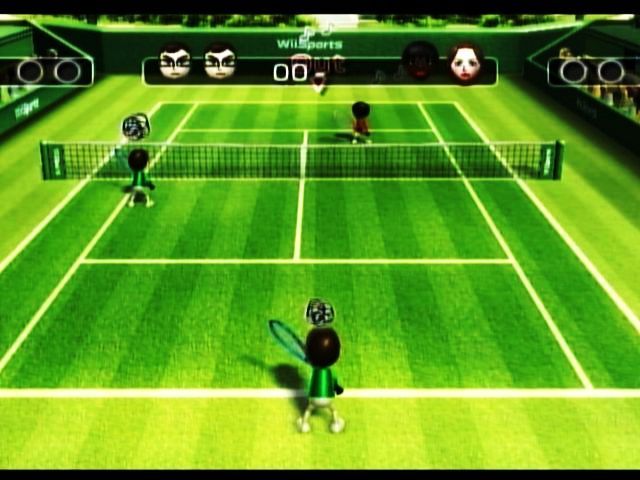}}  \hfill
    \subfloat[Mario Kart 8 - Wii U]{\includegraphics[width=0.19\textwidth]{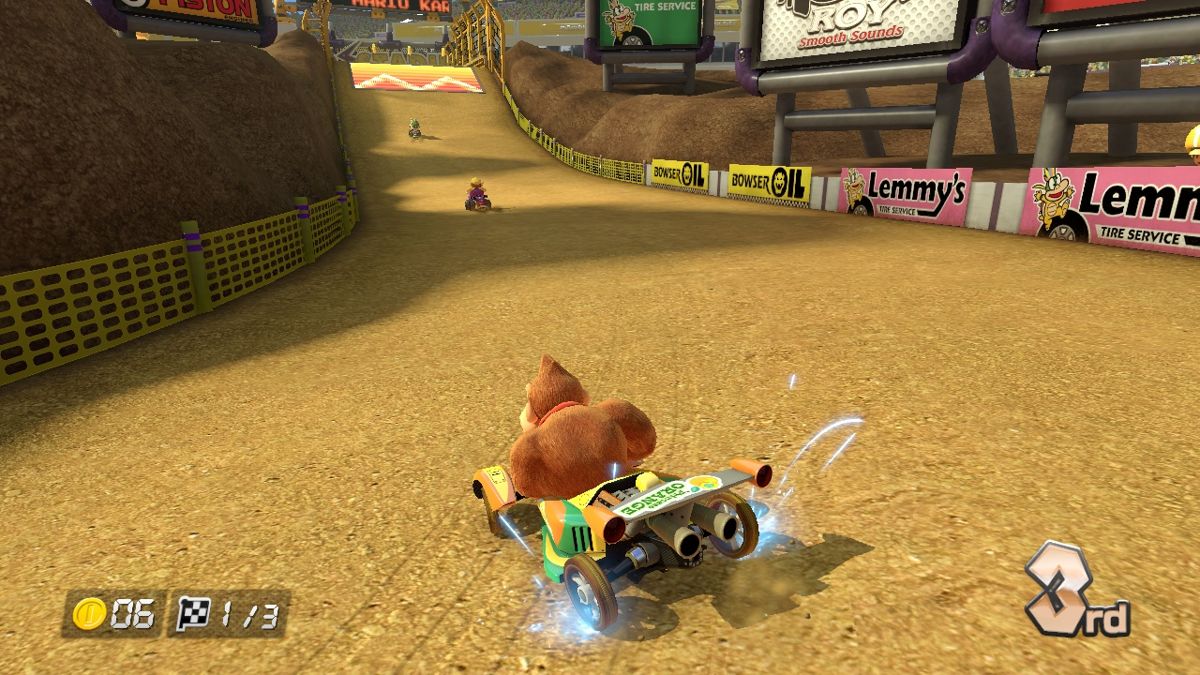}} \hfill
    \subfloat[The Witcher 3: Wild Hunt - PlayStation 4]{\includegraphics[width=0.19\textwidth]{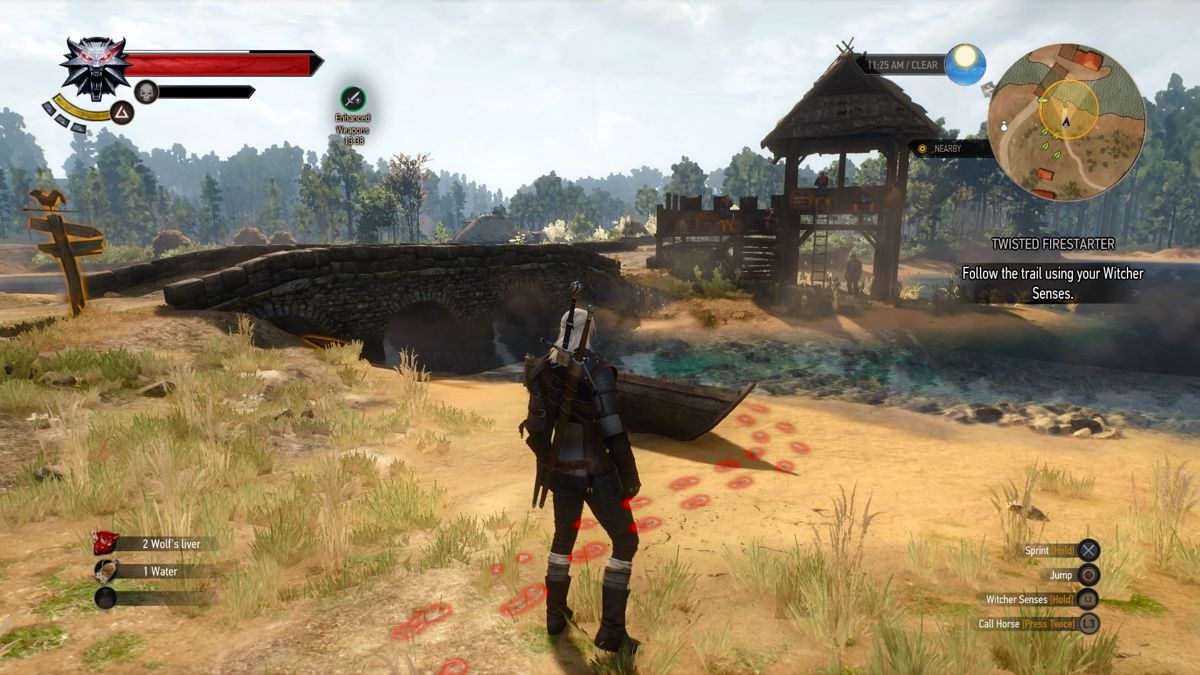}} 
    
    \subfloat[Horizon: Zero Dawn - PlayStation 4]{\includegraphics[width=0.19\textwidth]{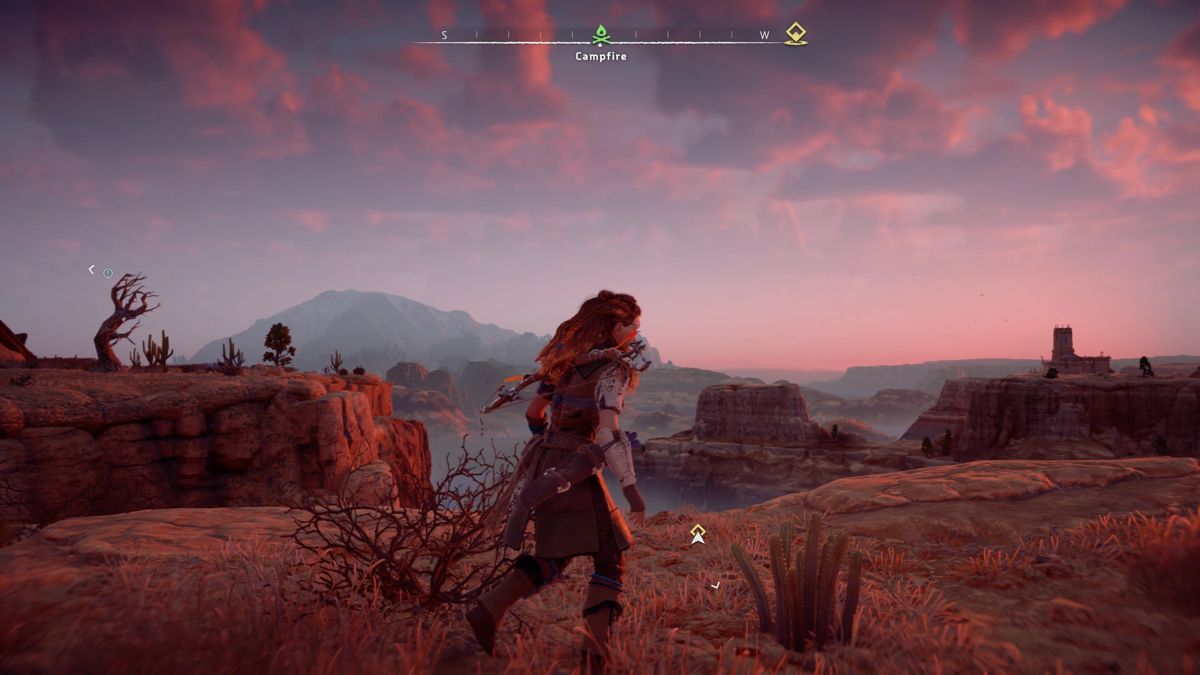}} \hfill
    \subfloat[Gears of War 4 - Xbox One]{\includegraphics[width=0.19\textwidth]{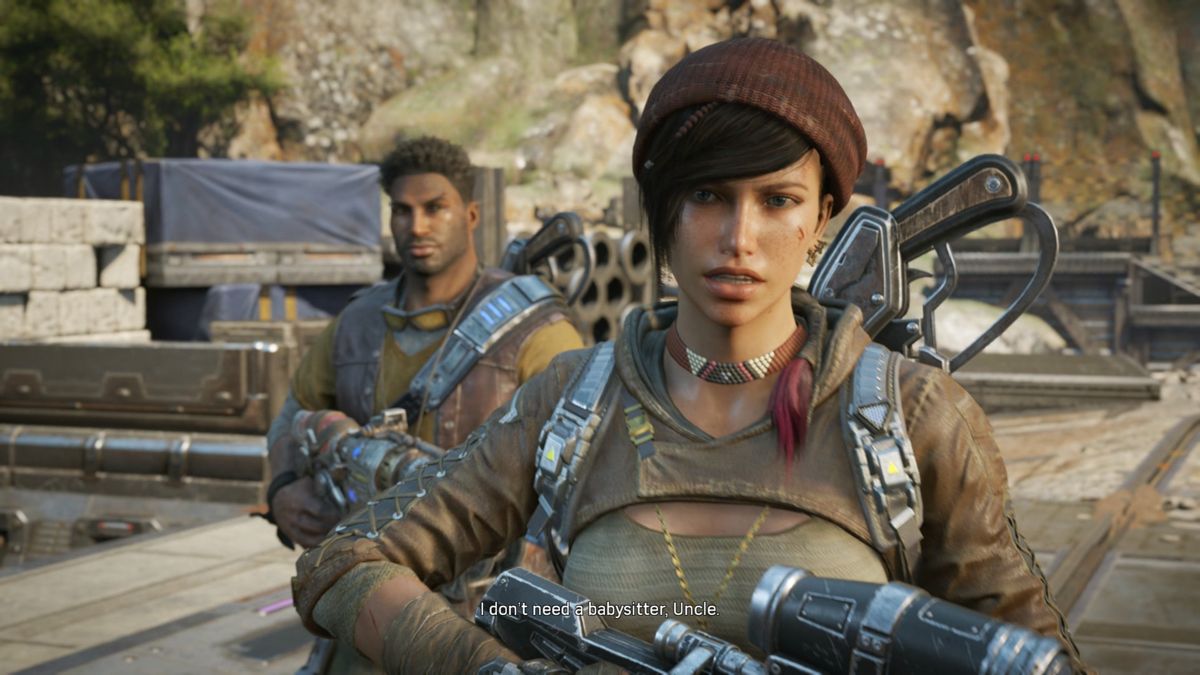}} \hfill
    \subfloat[The Legend of Zelda: Breath of the Wild - Nintendo Switch]{\includegraphics[width=0.19\textwidth]{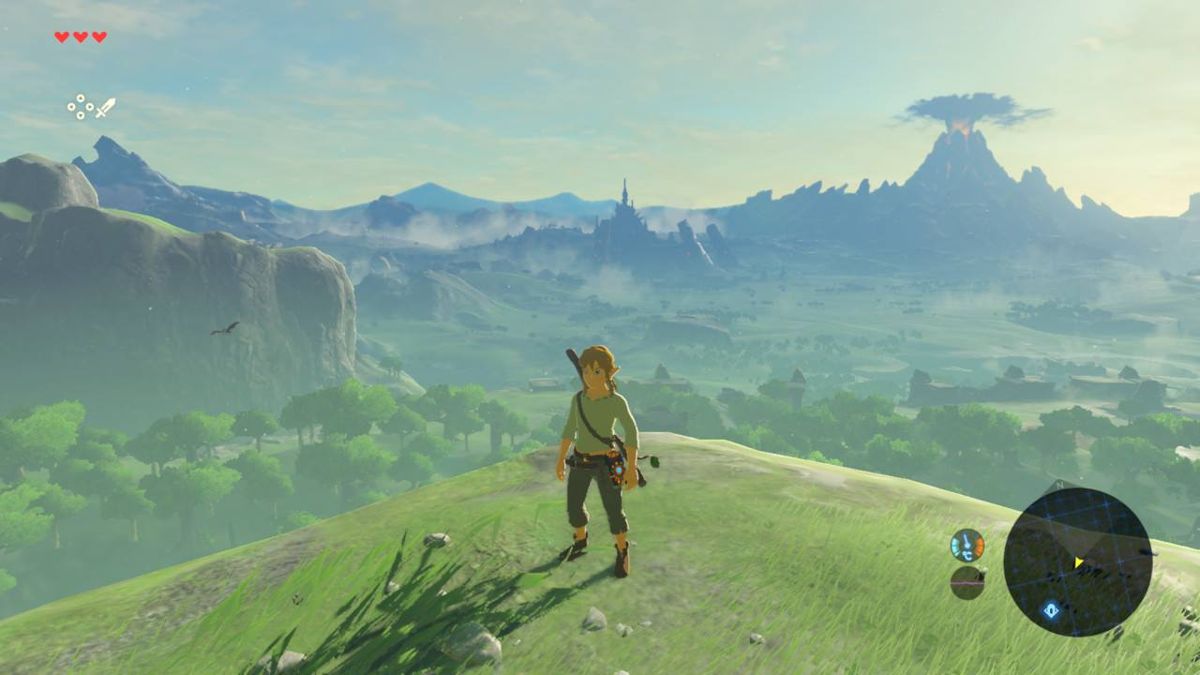}} \hfill
    \subfloat[Forza Horizon 5 - Xbox Series X/S]{\includegraphics[width=0.19\textwidth]{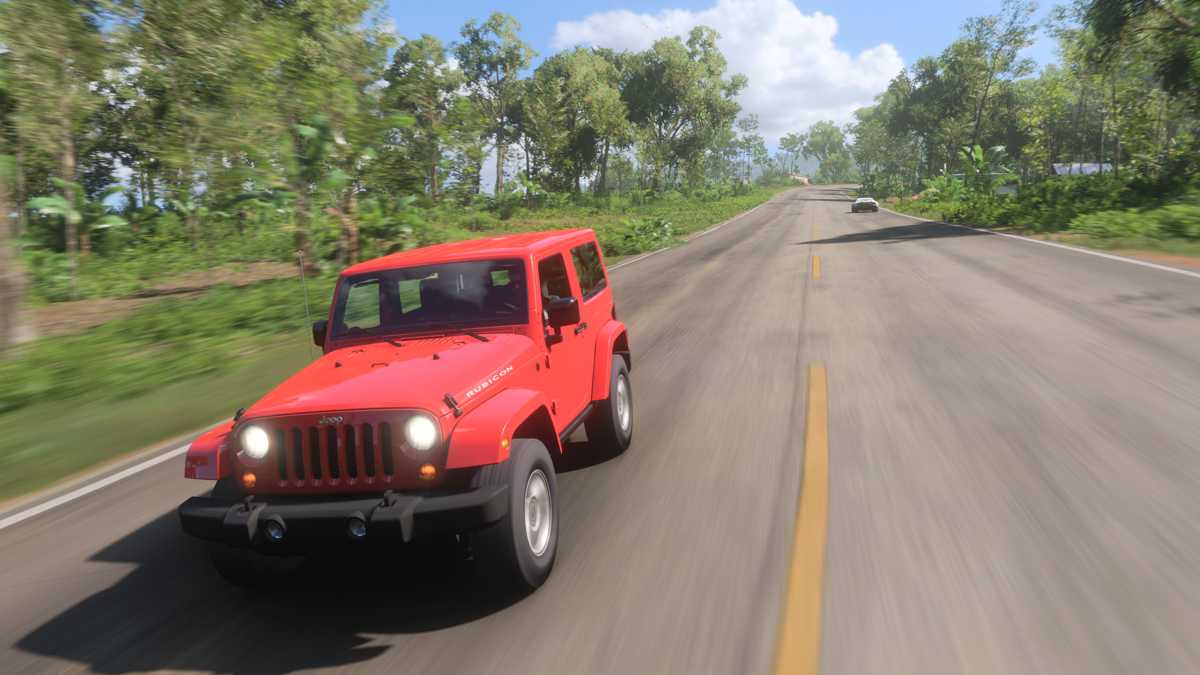}} \hfill
    \subfloat[Ratchet \& Clank: Rift Apart - PlayStation 5]{\includegraphics[width=0.19\textwidth]{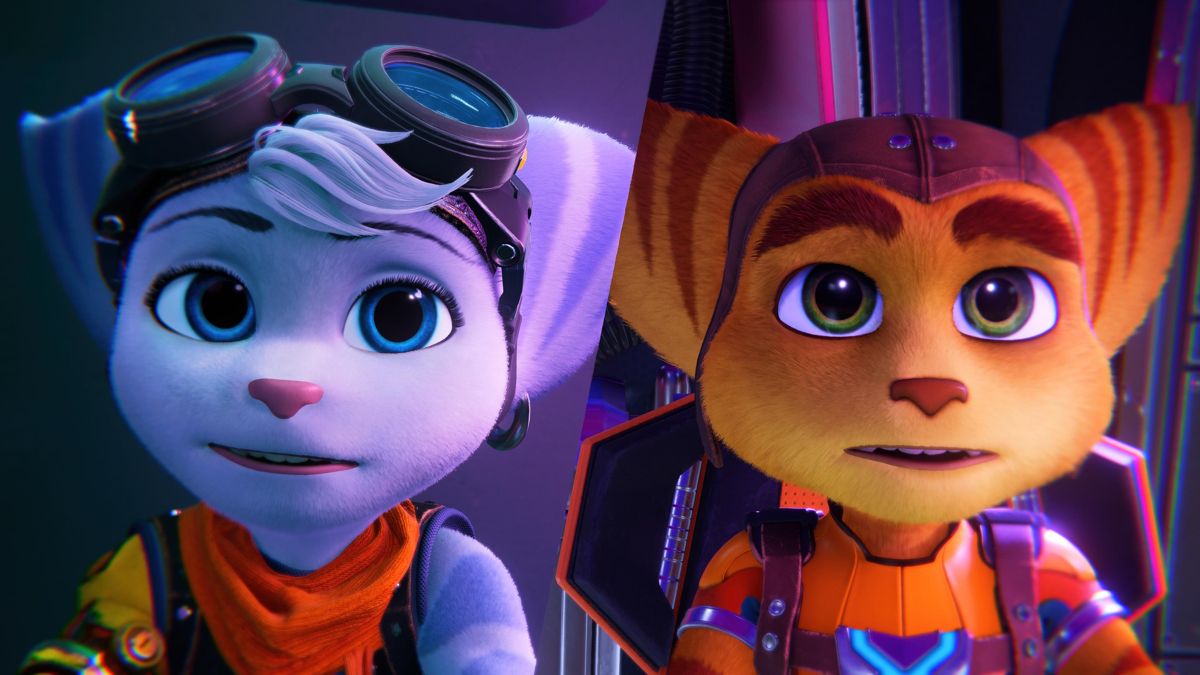}}
  
  \caption{Some examples of screenshots from the proposed dataset.}
  \label{fig:dataset}
  
\end{figure*}

Table \ref{tab:Dataset} provides an overview of the 22 chosen systems, detailing their total game and screenshot counts, as well as the specific number of games and screenshots selected to satisfy the ``at least 5 screenshots'' criterion. It is important to note that some of the newer systems, such as the Wii U, Nintendo Switch, Xbox Series X/S, and PlayStation 5, have fewer than a hundred games with available screenshots in Moby Games. Since the last three are current-generation consoles, it is expected that more screenshots will become available over time. However, this current limitation can restrict our analysis. Figure \ref{fig:dataset} shows some screenshots from the built dataset.

\begin{table*}
\caption{Home Console Video Game Systems, Their Manufacturers, and the Total/Selected Number of Games and Screenshots for the Study. Games with a minimum of five available screenshots in the database were chosen for analysis.}
\centering
\begin{tabular}{rccccc}
\toprule
\textbf{System} & \textbf{Manufacturer} & {\bf \makecell{Selected \\ Games}} & {\bf \makecell{Total \\ Games}} & {\bf \makecell{Selected \\ Screenshots}} & {\bf \makecell{Total \\ Screenshots}} \\
\midrule
Atari 2600 & Atari & 302   & 598   & 2148  & 2981 \\
NES   & Nintendo & 1236  & 1426  & 18277 & 18579 \\
Master System & Sega  & 322   & 357   & 4795  & 4850 \\
PC Engine & NEC/Hudson Soft & 213   & 276   & 3024  & 3099 \\
Mega Drive & Sega  & 926   & 1016  & 15838 & 15943 \\
Super Nintendo & Nintendo & 1113  & 1216  & 20926 & 21079 \\
Sega Saturn & Sega  & 249   & 809   & 4413  & 4536 \\
PlayStation & Sony  & 1197  & 2815  & 26831 & 27095 \\
Nintendo 64 & Nintendo & 172   & 378   & 2958  & 3125 \\
Dreamcast & Sega  & 155   & 553   & 2147  & 2217 \\
PlayStation 2 & Sony  & 677   & 3430  & 14157 & 14590 \\
GameCube & Nintendo & 149   & 621   & 3061  & 3094 \\
Xbox  & Microsoft & 161   & 1015  & 3351  & 3379 \\
Xbox 360 & Microsoft & 651   & 9357  & 9087  & 9426 \\
PlayStation 3 & Sony  & 255   & 15146 & 8358  & 8493 \\
Wii   & Nintendo & 118   & 2680  & 2519  & 2655 \\
Wii U & Nintendo & 31    & 1590  & 775   & 798 \\
PlayStation 4 & Sony  & 625   & 22733 & 22695 & 22912 \\
Xbox One & Microsoft & 119   & 16545 & 2397  & 2661 \\
Nintendo Switch & Nintendo & 77    & 12374 & 1736  & 1787 \\
Xbox Series X/S & Microsoft & 5     & 3256  & 37    & 44 \\
PlayStation 5 & Sony  & 43    & 2432  & 1351  & 1356 \\
\midrule
\textbf{Total} &       & \textbf{8796} & \textbf{100623} & \textbf{170881} & \textbf{174699} \\
\bottomrule
\end{tabular}%
\label{tab:Dataset}

\end{table*}

\captionsetup[subfloat]{labelformat=empty, justification=centering, font=footnotesize}

\section{CNN and Transformer Architectures}
\label{sec:CNNArchitectures}

In this section, the CNN and transformer architectures explored in this study are introduced, along with a description of the additional layers integrated to achieve successful screenshot classification. Table~\ref{tab:Models} offers an overview of the thirteen architectures under examination, highlighting their input image resolution, the size of their output in the last layer before the classification layer, the quantity of parameters involved, and citations to their respective references in the literature.

\begin{table*}

\caption{CNN and transformer architectures, some of their characteristics, and their references.}

\centering

\begin{tabular}{rcccl}

\toprule

{\bf Model} & {\bf \makecell{Input Image \\ Resolution}} & {\bf \makecell{Output of Last Layer \\ Before Classification}} & {\bf \makecell{Parameters}} & {\bf Reference} \\

\toprule

VGG16 &  $224 \times 224$ & $7 \times 7 \times 512$ &  $138.4$M & \cite{Simonyan2015} \\

ResNet50 &  $224 \times 224$ & $7 \times 7 \times 2048$ & $25.6$M & \cite{He2016} \\

ResNet152 &  $224 \times 224$ & $7 \times 7 \times 2048$ & $60.4$M & \cite{He2016} \\

MobileNet &  $224 \times 224$ & $7 \times 7 \times 1024$ &  $4.3$M & \cite{Howard2017} \\

DenseNet169 &  $224 \times 224$ & $7 \times 7 \times 1664$ & $14.3$M & \cite{Huang2017} \\

DenseNet201 &  $224 \times 224$ & $7 \times 7 \times 1920$ & $20.2$M & \cite{Huang2017} \\

EfficientNetB0 &  $224 \times 224$ & $7 \times 7 \times 1280$ &  $5.3$M & \cite{Tan2019} \\

EfficientNetB2 &  $260 \times 260$ & $9 \times 9 \times 1408$ &  $9.2$M & \cite{Tan2019} \\

EfficientNetB3 &  $300 \times 300$ & $10 \times 10 \times 1536$ &  $12.3$M & \cite{Tan2019} \\

EfficientNetV2S &  $384 \times 384$ & $12 \times 12 \times 1280$ &  $21.6$M & \cite{Tan2021} \\

ViT-B16 &  $224 \times 224$ & $768$ &  $8.66$M & \cite{Dosovitskiy2021} \\

ViT-L32 &  $384 \times 384$ & $1024$ &  $306.5$M & \cite{Dosovitskiy2021} \\

SwinT &  $224 \times 224$ & $7 \times 7 \times 768$ & $28.3$M & \cite{Liu2021} \\

\bottomrule

\end{tabular}

\label{tab:Models}

\end{table*}

\begin{figure*}
    \centering
    \includegraphics[width=\textwidth]{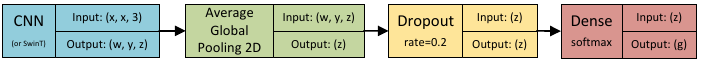}
    \caption{The proposed CNN and SwinT Transfer Learning architecture. The ViT architecture is similar but lacks the pooling layer because its output is one-dimensional.}
    \label{fig:CNN-Diagram}
\end{figure*}

Each of the thirteen network architectures is independently applied to each of the 22 home console datasets, resulting in a total of $286$ variations. The output from the last layer of the original CNN or Swin Transformer (SwinT), before the classification layer, is a 3-D tensor, which is then directed into a global average pooling layer. Subsequently, a dropout layer with a rate of $20\%$ is implemented to mitigate overfitting, followed by a large softmax classification layer with up to 1,236 outputs, corresponding to the number of game titles in the NES console dataset. This proposed architecture is visualized in Figure~\ref{fig:CNN-Diagram}, with  $x$ representing the dimensions of the input size (image size), $w$, $y$, and $z$ indicating the dimensions of the CNN or SwinT output in its final layer before classification (as detailed in Table~\ref{tab:Models}), and $g$ denoting the output layer’s dimensions, which depend on the number of games from the system being evaluated that are present in the dataset (as indicated in Table~\ref{tab:Dataset}). Note that the Visual Transformer (ViT) models produce 1-D tokens, therefore the pooling layer is not used in their case, and the outputs proceed directly to the dropout layer.

\section{Comparison}
\label{sec:CNNComparison}

This section presents experiments that compare the CNN and transformer models applied to the classification of screenshots from each of the 22 systems shown in Table \ref{tab:Dataset} individually. All experiments utilized Python and TensorFlow running on three distinct desktop computers equipped with four distinct NVIDIA GeForce GPU boards: GTX 970, GTX 1080, RTX 2060 SUPER, and RTX 4060 Ti\footnote{Access the source code at \url{https://github.com/fbreve/videogame}}.

For each CNN and transformer architecture, images were resized to meet the input size requirements of the respective models without maintaining the aspect ratio and were normalized according to each model's normalization function. No additional preprocessing was performed. The networks were initialized with pre-trained weights from the ImageNet dataset \cite{ILSVRC15}, which contains millions of images across hundreds of classes and is commonly used for transfer learning. These pre-trained weights are available with the corresponding CNN or transformer implementations. All experiments utilized Keras implementations of CNNs and transformers\footnote{CNNs: \url{https://keras.io/api/applications/}, ViT: \url{https://pypi.org/project/vit-keras/}, Swin: \url{https://pypi.org/project/tfswin/}}.

K-Fold Cross Validation, employing $k=5$, was applied universally across all datasets. Training utilized the Adam optimizer \cite{Kingma2014}, initiating with a learning rate of $10^{-3}$ and halving whenever the validation accuracy stagnated for 2 epochs, down to a minimum of $10^{-5}$. In each training stage, from the four folds composing the training subset, a random $20\%$ of images were allocated to the validation subset, ensuring consistent class proportions through stratification. All models underwent training for up to 50 epochs, with an early stopping criterion to cease training if the validation set loss failed to decrease during the last 10 epochs. Other hyperparameters were set to their TensorFlow defaults. These hyperparameters were chosen based on experiments conducted in prior studies using various types of images \cite{Breve2022,Breve2023}.

The results are detailed in Tables~\ref{tab:CNNResults}~and~\ref{tab:TransformerResults} for CNNs and transformers, respectively. Each result is the average from five different instances of each model, following the Cross Validation approach.

\begin{sidewaystable}

\caption{Accuracy achieved by the ten different CNN models in each of the $22$ screenshots datasets. Each model is executed five times following the Cross Validation approach. The highest accuracy for each dataset is highlighted in bold. Standard deviations are shown in parentheses.}

\centering

\resizebox{\textwidth}{!}{

\begin{tabular}{r|cc|cc|cc|cc|cc|cc|cc|cc|cc|cc|cc}
\textbf{System} & \multicolumn{2}{c|}{\textbf{VGG16}} & \multicolumn{2}{c|}{\textbf{ResNet50}} & \multicolumn{2}{c|}{\textbf{ResNet152}} & \multicolumn{2}{c|}{\textbf{MobileNet}} & \multicolumn{2}{c|}{\textbf{DenseNet169}} & \multicolumn{2}{c|}{\textbf{DenseNet201}} & \multicolumn{2}{c|}{\textbf{EfficientNetB0}} & \multicolumn{2}{c|}{\textbf{EfficientNetB2}} & \multicolumn{2}{c|}{\textbf{EfficientNetB3}} & \multicolumn{2}{c|}{\textbf{EfficientNetV2S}} & \multicolumn{2}{c}{\textbf{Average}} \\
\midrule
Atari 2600 & 48,47\% & (34,87\%) & 89,25\% & (0,58\%) & 87,90\% & (0,85\%) & 89,25\% & (1,16\%) & 89,39\% & (1,21\%) & 89,06\% & (0,53\%) & 88,92\% & (0,98\%) & 89,20\% & (1,39\%) & \textbf{90,36\%} & (1,52\%) & 88,92\% & (2,74\%) & 47,26\% & (44,50\%) \\
NES   & 50,87\% & (1,94\%) & 65,84\% & (2,30\%) & 65,65\% & (0,82\%) & 65,50\% & (3,87\%) & 67,19\% & (3,65\%) & 70,04\% & (1,91\%) & 68,45\% & (1,75\%) & 68,76\% & (3,27\%) & 67,80\% & (4,17\%) & \textbf{72,51\%} & (2,40\%) & 35,86\% & (33,11\%) \\
Master System & 21,46\% & (21,46\%) & 71,85\% & (0,17\%) & 66,80\% & (4,45\%) & 70,70\% & (0,77\%) & 74,52\% & (0,97\%) & 73,74\% & (1,14\%) & 71,24\% & (1,00\%) & 74,18\% & (0,30\%) & 74,29\% & (1,23\%) & \textbf{76,20\%} & (1,19\%) & 37,00\% & (35,81\%) \\
PC Engine & 30,55\% & (16,07\%) & 77,38\% & (1,26\%) & 75,36\% & (1,29\%) & 75,43\% & (1,05\%) & 79,89\% & (1,39\%) & 80,19\% & (0,70\%) & 77,71\% & (0,54\%) & 78,34\% & (0,83\%) & 78,90\% & (0,84\%) & \textbf{80,52\%} & (0,94\%) & 39,82\% & (38,07\%) \\
Mega Drive & 28,81\% & (23,31\%) & 68,43\% & (2,94\%) & 68,08\% & (2,00\%) & 67,51\% & (5,76\%) & 73,07\% & (2,51\%) & 73,44\% & (2,53\%) & 73,26\% & (0,43\%) & 72,46\% & (3,95\%) & 73,07\% & (4,22\%) & \textbf{74,56\%} & (3,22\%) & 37,74\% & (36,29\%) \\
Super Nintendo & 48,24\% & (1,70\%) & 68,23\% & (2,92\%) & 68,42\% & (1,29\%) & 67,24\% & (4,92\%) & 72,47\% & (3,46\%) & \textbf{74,51\%} & (0,80\%) & 70,15\% & (3,00\%) & 71,42\% & (3,25\%) & 70,81\% & (6,09\%) & 72,89\% & (1,72\%) & 37,23\% & (34,75\%) \\
Sega Saturn & 20,78\% & (13,98\%) & 70,04\% & (1,09\%) & 67,19\% & (1,40\%) & 69,73\% & (1,42\%) & 74,94\% & (2,11\%) & 75,10\% & (1,35\%) & 67,28\% & (1,53\%) & 68,41\% & (1,79\%) & 70,50\% & (0,98\%) & \textbf{75,91\%} & (1,11\%) & 35,80\% & (34,64\%) \\
PlayStation & 27,09\% & (21,75\%) & 67,41\% & (2,33\%) & 66,10\% & (2,43\%) & 69,08\% & (1,20\%) & 70,79\% & (2,78\%) & 71,19\% & (2,78\%) & 67,68\% & (4,59\%) & 72,27\% & (3,02\%) & 74,02\% & (1,03\%) & \textbf{74,77\%} & (2,49\%) & 36,85\% & (35,32\%) \\
Nintendo 64 & 23,88\% & (19,26\%) & 78,46\% & (1,56\%) & 76,10\% & (1,15\%) & 76,54\% & (2,06\%) & 83,27\% & (2,07\%) & \textbf{83,67\%} & (1,17\%) & 73,83\% & (1,14\%) & 76,81\% & (2,65\%) & 77,76\% & (0,90\%) & 82,28\% & (3,19\%) & 40,08\% & (38,73\%) \\
Dreamcast & 4,56\% & (2,25\%) & 70,47\% & (1,61\%) & 67,02\% & (0,99\%) & 65,21\% & (3,81\%) & 75,13\% & (1,26\%) & \textbf{75,36\%} & (1,90\%) & 63,49\% & (2,62\%) & 67,35\% & (3,00\%) & 67,58\% & (0,97\%) & 74,99\% & (1,83\%) & 33,74\% & (33,53\%) \\
PlayStation 2 & 28,17\% & (22,71\%) & 68,87\% & (0,87\%) & 66,25\% & (1,20\%) & 69,39\% & (0,83\%) & 72,73\% & (1,40\%) & 74,11\% & (0,39\%) & 68,36\% & (1,13\%) & 71,26\% & (2,42\%) & 73,45\% & (2,52\%) & \textbf{75,14\%} & (3,29\%) & 36,68\% & (35,17\%) \\
GameCube & 5,55\% & (2,92\%) & 77,49\% & (1,39\%) & 75,37\% & (0,80\%) & 74,16\% & (0,69\%) & 82,33\% & (0,55\%) & \textbf{82,88\%} & (1,14\%) & 72,07\% & (0,86\%) & 75,40\% & (3,22\%) & 74,98\% & (1,40\%) & 81,48\% & (2,30\%) & 37,16\% & (36,92\%) \\
Xbox  & 45,09\% & (19,18\%) & 79,80\% & (1,59\%) & 78,36\% & (0,87\%) & 78,04\% & (6,20\%) & 84,39\% & (1,85\%) & 84,03\% & (1,38\%) & 73,86\% & (2,62\%) & 79,29\% & (1,17\%) & 78,60\% & (2,33\%) & \textbf{84,45\%} & (1,32\%) & 42,14\% & (39,62\%) \\
Xbox 360 & 38,72\% & (17,06\%) & 74,72\% & (1,01\%) & 72,76\% & (1,15\%) & 74,13\% & (0,21\%) & 78,59\% & (0,79\%) & 79,43\% & (0,72\%) & 73,39\% & (0,67\%) & 76,23\% & (1,69\%) & 77,68\% & (0,86\%) & \textbf{80,25\%} & (1,17\%) & 39,35\% & (37,12\%) \\
PlayStation 3 & 43,80\% & (1,45\%) & 63,71\% & (1,32\%) & 62,91\% & (1,60\%) & 62,18\% & (5,40\%) & 70,73\% & (1,47\%) & 70,93\% & (0,90\%) & 66,58\% & (1,05\%) & 68,90\% & (1,12\%) & 71,82\% & (0,92\%) & \textbf{71,98\%} & (3,04\%) & 35,05\% & (32,53\%) \\
Wii   & 20,06\% & (17,62\%) & 76,35\% & (1,54\%) & 75,82\% & (1,49\%) & 76,38\% & (0,75\%) & 83,37\% & (0,90\%) & 83,05\% & (0,79\%) & 75,55\% & (3,02\%) & 76,74\% & (1,90\%) & 78,40\% & (1,64\%) & \textbf{83,96\%} & (0,69\%) & 39,63\% & (38,55\%) \\
Wii U & 14,71\% & (1,44\%) & 67,69\% & (5,09\%) & 66,06\% & (3,23\%) & 65,94\% & (1,75\%) & 72,64\% & (4,73\%) & 73,03\% & (2,66\%) & 69,42\% & (1,66\%) & 70,06\% & (4,44\%) & 71,87\% & (2,93\%) & \textbf{74,97\%} & (2,43\%) & 35,08\% & (34,39\%) \\
PlayStation 4 & 36,91\% & (1,27\%) & 52,17\% & (5,04\%) & 54,38\% & (2,11\%) & 56,01\% & (0,63\%) & 61,51\% & (0,41\%) & 62,19\% & (0,70\%) & 59,75\% & (1,77\%) & 62,73\% & (0,98\%) & \textbf{64,77\%} & (1,53\%) & 63,17\% & (2,88\%) & 30,78\% & (28,70\%) \\
Xbox One & 25,99\% & (4,44\%) & 61,37\% & (1,30\%) & 58,62\% & (1,03\%) & 55,82\% & (1,84\%) & 64,58\% & (1,08\%) & 64,96\% & (1,33\%) & 57,91\% & (1,96\%) & 58,58\% & (3,06\%) & 63,50\% & (2,92\%) & \textbf{66,25\%} & (2,17\%) & 31,02\% & (29,66\%) \\
Nintendo Switch & 17,57\% & (2,95\%) & 75,63\% & (2,30\%) & 70,28\% & (1,85\%) & 73,62\% & (1,09\%) & 80,01\% & (1,76\%) & 79,90\% & (1,55\%) & 75,11\% & (0,60\%) & 78,11\% & (1,34\%) & 78,86\% & (1,34\%) & \textbf{80,47\%} & (1,70\%) & 37,80\% & (36,84\%) \\
Xbox Series X/S & 38,21\% & (6,93\%) & 43,93\% & (12,25\%) & 27,14\% & (15,75\%) & 76,43\% & (16,54\%) & 60,71\% & (21,64\%) & 78,57\% & (5,98\%) & 91,78\% & (6,74\%) & 97,14\% & (5,71\%) & 97,14\% & (5,71\%) & \textbf{100,00\%} & (0,00\%) & 41,33\% & (39,42\%) \\
PlayStation 5 & 14,95\% & (3,03\%) & 59,81\% & (1,06\%) & 54,25\% & (6,38\%) & 57,14\% & (1,63\%) & 65,80\% & (1,83\%) & 67,58\% & (2,22\%) & 58,63\% & (3,82\%) & 61,88\% & (2,72\%) & 62,99\% & (2,57\%) & \textbf{68,10\%} & (2,52\%) & 30,93\% & (30,20\%) \\
\midrule
\textbf{Average} & 28,84\% & (11,71\%) & 69,50\% & (2,34\%) & 66,86\% & (2,46\%) & 69,79\% & (2,89\%) & 74,46\% & (2,72\%) & 75,77\% & (1,57\%) & 71,11\% & (1,98\%) & 73,43\% & (2,42\%) & 74,51\% & (2,21\%) & \textbf{77,44\%} & (2,02\%) & 37,20\% & (35,63\%) \\
\end{tabular}%

}

\label{tab:CNNResults}
\end{sidewaystable}

Regarding the CNNs, EfficientNetV2S achieved the best accuracy in 16 of the 22 systems, as well as the best average accuracy across all systems ($77.44\%$). Regarding the systems, the best accuracy is achieved with the \emph{Xbox Series X/S} using EfficientNetV2S ($100\%$). However, it is worth noticing that this system only had 37 screenshots from a total of five games. The \emph{Atari 2600} is the system with the best average accuracy ($84.64\%$) among all tested models. This is likely related to the simpler graphics of this second-generation console compared to newer systems. Most games for the \emph{Atari 2600} do not exhibit significant screen variation. On the other hand, newer systems with complex graphics like \emph{PlayStation 4}, \emph{PlayStation 5} and \emph{XBox One} are among those with the lowest average accuracy.

In simpler tasks, smaller architectures often perform as well as larger ones. It's valuable to consider these smaller networks when selecting the optimal architecture for a task because if a smaller network can yield comparable results in less computational time, there is no justification for employing a larger one. This rationale led to the inclusion of MobileNet and EfficientNetB0 in this comparison. However, in the context of video game detection by screenshot, it became evident that larger networks like DenseNet and the larger versions of EfficientNet outperformed the smaller ones.  Therefore, it is not possible to use smaller networks without compromising accuracy.

Earlier architectures such as VGG and ResNet performed worse compared to newer models like DenseNet and EfficientNet. Notably, VGG struggled to learn the training subset in many instances, as indicated by its significantly lower accuracy and higher standard deviation relative to the other models.

Over recent years, transformer architectures have begun to outperform CNN models in many image classification tasks where CNNs once dominated. However, this is not the case for the video game identification by screenshots task. The three transformer architectures evaluated performed worse than all the CNN models except for VGG16. ViT-B16 achieved the highest average accuracy among the transformers, but only at $51.19\%$. SwinT was the best among transformers in nine of the 22 systems, though its overall average was lower at $44.06\%$. In many instances, the transformers struggled to learn the training subset, contributing to their higher standard deviation. Even when they did learn, their accuracy remained substandard compared to CNNs. It is known that transformers require larger training sets than CNNs to outperform them in most scenarios. This is likely the case for this task, where the number of screenshots per game (individuals per class) is relatively low.

\begin{table*}

\caption{Accuracy achieved by the three different transformer models in each of the $22$ screenshots datasets. Each model is executed five times following the Cross Validation approach. The highest accuracy for each dataset is highlighted in bold. Standard deviations are shown in parentheses.}

\centering

\begin{tabular}{r|cc|cc|cc|cc}
\textbf{System} & \multicolumn{2}{c|}{\textbf{ViT-B16}} & \multicolumn{2}{c|}{\textbf{ViT-L32}} & \multicolumn{2}{c|}{\textbf{SwinT}} & \multicolumn{2}{c}{\textbf{Average}} \\
\midrule
Atari 2600 & 72,49\% & (2,49\%) & \textbf{74,30\%} & (2,88\%) & 2,19\% & (0,56\%) & 30,87\% & (16,48\%) \\
NES   & 51,06\% & (7,03\%) & \textbf{58,17\%} & (1,94\%) & 54,53\% & (1,38\%) & 51,06\% & (24,61\%) \\
Master System & 55,91\% & (5,25\%) & \textbf{58,06\%} & (9,13\%) & 40,58\% & (25,99\%) & 55,91\% & (27,80\%) \\
PC Engine & 69,74\% & (1,21\%) & 57,05\% & (11,06\%) & \textbf{73,21\%} & (3,26\%) & 69,74\% & (29,16\%) \\
Mega Drive & 52,43\% & (8,28\%) & \textbf{55,84\%} & (10,09\%) & 35,07\% & (28,40\%) & 52,43\% & (27,54\%) \\
Super Nintendo & 53,12\% & (8,57\%) & \textbf{53,58\%} & (11,08\%) & 33,66\% & (32,60\%) & 53,12\% & (27,90\%) \\
Sega Saturn & 52,07\% & (5,51\%) & \textbf{52,96\%} & (2,73\%) & 50,63\% & (7,15\%) & 52,07\% & (23,80\%) \\
PlayStation & 42,89\% & (8,96\%) & 48,74\% & (8,98\%) & \textbf{54,70\%} & (3,23\%) & 42,89\% & (24,92\%) \\
Nintendo 64 & 57,03\% & (3,24\%) & 49,03\% & (18,10\%) & \textbf{57,97\%} & (25,75\%) & 57,03\% & (30,82\%) \\
Dreamcast & 51,61\% & (2,58\%) & 50,02\% & (2,62\%) & \textbf{57,40\%} & (8,27\%) & 51,61\% & (24,18\%) \\
PlayStation 2 & \textbf{51,08\%} & (3,23\%) & 47,83\% & (8,04\%) & 44,60\% & (22,13\%) & 51,08\% & (25,17\%) \\
GameCube & \textbf{57,76\%} & (5,47\%) & 46,40\% & (20,30\%) & 22,41\% & (24,37\%) & 57,76\% & (23,79\%) \\
Xbox  & 52,77\% & (9,66\%) & 52,28\% & (4,49\%) & \textbf{55,06\%} & (26,55\%) & 52,77\% & (29,61\%) \\
Xbox 360 & 42,65\% & (18,69\%) & 46,77\% & (18,25\%) & \textbf{59,32\%} & (2,59\%) & 42,65\% & (29,12\%) \\
PlayStation 3 & 45,32\% & (2,40\%) & 33,02\% & (10,75\%) & \textbf{46,53\%} & (2,65\%) & 45,32\% & (19,07\%) \\
Wii   & \textbf{60,66\%} & (5,14\%) & 48,40\% & (20,21\%) & 43,57\% & (32,85\%) & 60,66\% & (30,03\%) \\
Wii U & \textbf{50,19\%} & (3,35\%) & 49,94\% & (4,25\%) & 33,08\% & (21,20\%) & 50,19\% & (22,36\%) \\
PlayStation 4 & 38,25\% & (6,17\%) & 28,60\% & (10,49\%) & \textbf{46,78\%} & (1,69\%) & 38,25\% & (18,75\%) \\
Xbox One & \textbf{34,88\%} & (10,77\%) & 32,30\% & (13,86\%) & 33,92\% & (17,75\%) & 34,88\% & (21,72\%) \\
Nintendo Switch & 50,57\% & (7,22\%) & \textbf{53,68\%} & (5,90\%) & 40,23\% & (22,26\%) & 50,57\% & (25,86\%) \\
Xbox Series X/S & 52,50\% & (22,52\%) & 48,93\% & (7,79\%) & \textbf{58,21\%} & (22,97\%) & 52,50\% & (32,08\%) \\
PlayStation 5 & \textbf{31,17\%} & (8,47\%) & 27,09\% & (1,99\%) & 25,57\% & (11,73\%) & 31,17\% & (14,97\%) \\
\midrule
\textbf{Average} & \textbf{51,19\%} & (7,10\%) & 48,77\% & (9,32\%) & 44,06\% & (15,70\%) & 49,30\% & (24,99\%) \\
\end{tabular}%

\label{tab:TransformerResults}
\end{table*}

\section{Alternative Initial Weights}
\label{sec:AlternativeWeights}

The ImageNet weights are commonly used in many transfer learning scenarios with success. Through fine-tuning, these weights can be adapted to perform many different tasks. However, it is expected that transferring weights from a similar task might enhance accuracy and reduce training times compared to using the ImageNet weights. Hence, investigating whether this holds true for the game identification by screenshots task is worthwhile.

To conduct these experiments, we used the platform listed on MobyGames as `Arcade'. They have included all the arcade games in this single platform. Therefore, it contains games contemporary to multiple home console generations, with graphics of increasing complexity being released over the years. The \emph{Arcade} screenshots were obtained using the same criteria applied in sourcing screenshots from home console systems. Out of 3,125 games and 24,714 screenshots, 1,633 games and 24,235 screenshots were selected based on the criterion of `at least five screenshots per game.'  It is worth noticing that some arcade games were later ported to home consoles, usually with simpler graphics due to hardware limitations. Despite that, convolutional layers trained in similar graphics could be more easily fine-tuned to detect the home version counterparts. 

The three architectures that demonstrated the best accuracies in the previous section — DenseNet201, EfficientNetB3, and EfficientNetV2S — were chosen for application in these new experiments. Each of them was trained using the entire \emph{Arcade} dataset, utilizing identical parameters as outlined earlier. The weights obtained from training on the \emph{Arcade} dataset were subsequently employed as initial weights for training these architectures with screenshots from each of the 22 home console systems. It is worth noting that the golden age of arcades spanned from the late 1970s to the early 1980s, and the number of new releases declined significantly afterward. In the new millennium, there are even fewer arcade game releases. Consequently, it is expected that earlier home consoles could derive greater benefit from using the \emph{Arcade} weights.

Tables~\ref{tab:ArcadeWeightsDenseNet201}, \ref{tab:ArcadeWeightsEfficientNetB3}, and \ref{tab:ArcadeWeightsEfficientNetV2S} display the accuracy achieved and the epochs required to train each network, using random weights initialization and both the ImageNet and \emph{Arcade} weights. In all scenarios, random weights resulted in lower accuracy and required more epochs for convergence compared to the other weight initialization methods. Conversely, Arcade weights outperformed ImageNet weights in most scenarios, though not all. For DenseNet201 (Table \ref{tab:ArcadeWeightsDenseNet201}), employing the \emph{Arcade} weights resulted in improved accuracy for only 10 out of the 22 systems. However, training times decreased for 20 of the 22 systems. Overall, while the average accuracy remained the same ($75.77\%$), the average number of epochs needed to train the network decreased from $24.6$ to $22.0$.

\begin{table*}

\caption{Accuracy and epochs required to train the DenseNet201 architecture across 22 screenshot datasets, using random weight initialization, ImageNet weights, and Arcade dataset weights as initial weights. Each model is executed five times using the Cross Validation approach. The highest accuracy and the lowest number of epochs for each dataset are highlighted in bold. Standard deviations are shown in parentheses.}

\centering

\begin{tabular}{r|cccc|cccc|cccc}
\toprule
\textbf{Weights} & \multicolumn{4}{c|}{\textbf{Random}} & \multicolumn{4}{c|}{\textbf{ImageNet}} & \multicolumn{4}{c}{\textbf{Arcade}} \\
\textbf{System} & \multicolumn{2}{c}{\textbf{Epochs}} & \multicolumn{2}{c|}{\textbf{Accuracy}} & \multicolumn{2}{c}{\textbf{Epochs}} & \multicolumn{2}{c|}{\textbf{Accuracy}} & \multicolumn{2}{c}{\textbf{Epochs}} & \multicolumn{2}{c}{\textbf{Accuracy}} \\
\midrule
Atari 2600 & 44.6  & (3.2) & 80.82\% & (2.17\%) & 23.6  & (1.5) & 89.06\% & (0.53\%) & \textbf{17.6} & (0.8) & \textbf{89.99\%} & (0.98\%) \\
NES   & 25.4  & (1.2) & 65.46\% & (0.85\%) & 23.0  & (3.9) & 70.04\% & (1.91\%) & \textbf{20.8} & (2.5) & \textbf{70.77\%} & (3.14\%) \\
Master System & 38.2  & (3.0) & 67.11\% & (1.32\%) & 24.0  & (1.5) & 73.74\% & (1.14\%) & \textbf{20.0} & (0.9) & \textbf{74.14\%} & (1.00\%) \\
PC Engine & 38.2  & (5.1) & 74.90\% & (0.68\%) & 23.6  & (2.1) & 80.19\% & (0.70\%) & \textbf{18.4} & (1.4) & \textbf{80.52\%} & (1.59\%) \\
Mega Drive & 29.2  & (2.9) & 67.40\% & (1.74\%) & 24.0  & (3.0) & \textbf{73.44\%} & (2.53\%) & \textbf{21.4} & (4.1) & 71.48\% & (4.19\%) \\
Super Nintendo & 26.8  & (2.1) & 66.48\% & (1.03\%) & 25.8  & (0.7) & \textbf{74.51\%} & (0.80\%) & \textbf{22.0} & (2.9) & 73.32\% & (2.36\%) \\
Sega Saturn & 37.8  & (3.9) & 65.17\% & (2.42\%) & 22.4  & (1.5) & 75.10\% & (1.35\%) & \textbf{19.2} & (1.3) & \textbf{83.14\%} & (1.15\%) \\
PlayStation & 25.4  & (1.7) & 63.52\% & (2.36\%) & \textbf{22.6} & (3.0) & 71.19\% & (2.78\%) & 23.0  & (3.8) & \textbf{71.59\%} & (2.44\%) \\
Nintendo 64 & 46.0  & (3.6) & 73.93\% & (3.35\%) & 26.2  & (3.0) & \textbf{83.67\%} & (1.17\%) & \textbf{19.2} & (0.7) & 83.37\% & (2.11\%) \\
Dreamcast & 48.8  & (1.2) & 63.35\% & (2.18\%) & 25.0  & (2.8) & 75.36\% & (1.90\%) & \textbf{20.4} & (3.0) & \textbf{76.06\%} & (1.40\%) \\
PlayStation 2 & 37.8  & (5.5) & 62.00\% & (2.22\%) & 25.2  & (3.2) & \textbf{74.11\%} & (0.39\%) & \textbf{22.0} & (2.0) & 73.13\% & (0.47\%) \\
GameCube & 50.0  & (0.0) & 71.71\% & (1.22\%) & 23.6  & (1.5) & \textbf{82.88\%} & (1.14\%) & \textbf{22.8} & (2.3) & 82.52\% & (1.13\%) \\
Xbox  & 46.2  & (4.0) & 74.25\% & (1.05\%) & 24.0  & (1.4) & 84.03\% & (1.38\%) & \textbf{22.4} & (1.4) & \textbf{84.06\%} & (1.62\%) \\
Xbox 360 & 41.8  & (5.2) & 68.01\% & (1.03\%) & 23.6  & (2.4) & \textbf{79.43\%} & (0.72\%) & \textbf{21.6} & (1.5) & 78.71\% & (0.54\%) \\
PlayStation 3 & 36.8  & (8.5) & 52.08\% & (6.77\%) & 23.8  & (2.3) & 70.93\% & (0.90\%) & \textbf{21.0} & (1.4) & \textbf{70.96\%} & (0.54\%) \\
Wii   & 46.6  & (3.6) & 73.96\% & (1.60\%) & 23.4  & (2.1) & \textbf{83.05\%} & (0.79\%) & \textbf{20.6} & (1.4) & 82.49\% & (0.71\%) \\
Wii U & 50.0  & (0.0) & 56.39\% & (2.02\%) & 25.6  & (6.4) & \textbf{73.03\%} & (2.66\%) & \textbf{20.6} & (2.7) & 72.90\% & (3.46\%) \\
PlayStation 4 & 26.6  & (4.6) & 49.18\% & (5.30\%) & 24.0  & (1.9) & \textbf{62.19\%} & (0.70\%) & \textbf{22.4} & (1.5) & 61.63\% & (0.55\%) \\
Xbox One & 43.6  & (1.9) & 54.28\% & (1.27\%) & 21.4  & (2.6) & 64.96\% & (1.33\%) & \textbf{20.2} & (2.4) & \textbf{66.25\%} & (1.41\%) \\
Nintendo Switch & 49.2  & (1.6) & 67.74\% & (2.49\%) & 24.8  & (4.3) & \textbf{79.90\%} & (1.55\%) & \textbf{24.2} & (4.3) & 78.98\% & (0.61\%) \\
Xbox Series X/S & 20.6  & (15.0) & 32.14\% & (12.78\%) & \textbf{36.2} & (16.9) & \textbf{78.57\%} & (5.98\%) & 42.4  & (15.2) & 73.57\% & (13.80\%) \\
PlayStation 5 & 42.4  & (2.7) & 47.82\% & (1.97\%) & 24.6  & (4.0) & \textbf{67.58\%} & (2.22\%) & \textbf{21.8} & (2.6) & 67.36\% & (1.29\%) \\
\midrule
\textbf{Average} & 38.7  & (3.7) & 63.53\% & (2.63\%) & 24.6  & (3.3) & \textbf{75.77\%} & (1.57\%) & \textbf{22.0} & (2.7) & \textbf{75.77\%} & (2.11\%) \\
\bottomrule
\end{tabular}%

\label{tab:ArcadeWeightsDenseNet201}
\end{table*}

For EfficientNetB3 (Table~\ref{tab:ArcadeWeightsEfficientNetB3}), employing the \emph{Arcade} weights led to improved accuracy for 19 of the 22 systems. Moreover, training times decreased for 20 of the 22 systems. Overall, the average accuracy increased from $74.51\%$ to $76.36\%$, while the average number of epochs required to train the network decreased from $23.7$ to $20.5$. This architecture demonstrated the most significant improvement using the Arcade weights.

\begin{table*}

\caption{Accuracy and epochs required to train the EfficientNetB3 architecture across 22 screenshot datasets, using random weight initialization, ImageNet weights, and Arcade dataset weights as initial weights. Each model is executed five times using the Cross Validation approach. The highest accuracy and the lowest number of epochs for each dataset are highlighted in bold. Standard deviations are shown in parentheses.}

\centering

\begin{tabular}{r|cccc|cccc|cccc}
\toprule
\textbf{Weights} & \multicolumn{4}{c|}{\textbf{Random}} & \multicolumn{4}{c|}{\textbf{ImageNet}} & \multicolumn{4}{c}{\textbf{Arcade}} \\
\textbf{System} & \multicolumn{2}{c}{\textbf{Epochs}} & \multicolumn{2}{c|}{\textbf{Accuracy}} & \multicolumn{2}{c}{\textbf{Epochs}} & \multicolumn{2}{c|}{\textbf{Accuracy}} & \multicolumn{2}{c}{\textbf{Epochs}} & \multicolumn{2}{c}{\textbf{Accuracy}} \\
Atari 2600 & 47.4  & (3.6) & 77.47\% & (2.28\%) & 28.4  & (6.1) & 90.36\% & (1.52\%) & \textbf{24.8} & (6.7) & \textbf{90.64\%} & (1.25\%) \\
NES   & 22.0  & (0.9) & 53.75\% & (2.20\%) & 15.2  & (1.5) & 67.80\% & (4.17\%) & \textbf{13.6} & (0.5) & \textbf{72.14\%} & (0.74\%) \\
Master System & 22.0  & (1.1) & 48.61\% & (3.19\%) & 22.0  & (1.7) & 74.29\% & (1.23\%) & \textbf{16.2} & (2.7) & \textbf{75.62\%} & (0.97\%) \\
PC Engine & 29.0  & (3.2) & 64.42\% & (1.95\%) & 19.2  & (1.0) & 78.90\% & (0.84\%) & \textbf{15.4} & (0.5) & \textbf{81.08\%} & (0.71\%) \\
Mega Drive & 21.2  & (0.4) & 54.28\% & (2.34\%) & 18.6  & (2.9) & 73.07\% & (4.22\%) & \textbf{13.8} & (0.8) & \textbf{74.20\%} & (0.60\%) \\
Super Nintendo & 21.8  & (1.2) & 52.30\% & (1.14\%) & 18.2  & (3.7) & 70.81\% & (6.09\%) & \textbf{15.4} & (2.6) & \textbf{73.95\%} & (1.53\%) \\
Sega Saturn & 22.6  & (3.2) & 45.93\% & (5.07\%) & 22.8  & (2.8) & 70.50\% & (0.98\%) & \textbf{15.2} & (0.8) & \textbf{75.10\%} & (1.60\%) \\
PlayStation & 22.2  & (0.4) & 53.81\% & (1.52\%) & 21.2  & (3.1) & \textbf{74.02\%} & (1.03\%) & \textbf{15.4} & (1.7) & 73.56\% & (0.63\%) \\
Nintendo 64 & 34.0  & (10.5) & 58.18\% & (1.26\%) & 31.6  & (10.3) & 77.76\% & (0.90\%) & \textbf{23.8} & (5.6) & \textbf{83.23\%} & (1.66\%) \\
Dreamcast & 35.0  & (6.8) & 48.44\% & (2.78\%) & 23.0  & (1.7) & 67.58\% & (0.97\%) & \textbf{21.6} & (2.2) & \textbf{76.43\%} & (1.52\%) \\
PlayStation 2 & 23.4  & (0.8) & 52.47\% & (1.46\%) & 21.6  & (3.3) & 73.45\% & (2.52\%) & \textbf{19.4} & (2.0) & \textbf{75.31\%} & (1.21\%) \\
GameCube & 30.2  & (4.0) & 58.80\% & (1.22\%) & 24.2  & (2.9) & 74.98\% & (1.40\%) & \textbf{23.8} & (6.0) & \textbf{83.96\%} & (0.96\%) \\
Xbox  & 27.6  & (2.6) & 63.89\% & (1.43\%) & 29.8  & (5.7) & 78.60\% & (2.33\%) & \textbf{21.8} & (3.7) & \textbf{85.11\%} & (1.50\%) \\
Xbox 360 & 25.6  & (3.8) & 52.29\% & (3.92\%) & 20.8  & (0.8) & 77.68\% & (0.86\%) & \textbf{20.2} & (1.6) & \textbf{79.55\%} & (0.65\%) \\
PlayStation 3 & 24.0  & (0.9) & 45.97\% & (1.19\%) & 20.6  & (3.1) & 71.82\% & (0.92\%) & \textbf{18.0} & (2.1) & \textbf{70.88\%} & (0.47\%) \\
Wii   & 35.6  & (6.7) & 62.13\% & (1.53\%) & 22.6  & (1.4) & 78.40\% & (1.64\%) & \textbf{18.4} & (4.4) & \textbf{81.70\%} & (1.11\%) \\
Wii U & 33.2  & (17.6) & 27.48\% & (16.66\%) & \textbf{22.0} & (4.2) & 71.87\% & (2.93\%) & 28.2  & (4.8) & \textbf{76.52\%} & (4.23\%) \\
PlayStation 4 & 22.2  & (1.7) & 40.53\% & (0.99\%) & 20.4  & (1.6) & \textbf{64.77\%} & (1.53\%) & \textbf{14.8} & (1.6) & 60.60\% & (1.69\%) \\
Xbox One & 22.4  & (1.0) & 40.76\% & (1.17\%) & 24.4  & (1.4) & 63.50\% & (2.92\%) & \textbf{15.2} & (1.5) & \textbf{65.96\%} & (1.38\%) \\
Nintendo Switch & 27.4  & (2.7) & 44.24\% & (1.35\%) & 32.4  & (7.2) & 78.86\% & (1.34\%) & \textbf{22.4} & (3.4) & \textbf{81.80\%} & (1.09\%) \\
Xbox Series X/S & 47.6  & (4.8) & 16.07\% & (4.52\%) & \textbf{40.0} & (13.1) & \textbf{97.14\%} & (5.71\%) & 50.0  & (0.0) & 75.71\% & (16.66\%) \\
PlayStation 5 & 24.6  & (1.5) & 35.75\% & (3.38\%) & 23.0  & (1.1) & 62.99\% & (2.57\%) & \textbf{22.8} & (5.8) & \textbf{66.76\%} & (1.68\%) \\
\midrule
\textbf{Average} & 28.2  & (3.6) & 49.89\% & (2.84\%) & 23.7  & (3.7) & 74.51\% & (2.21\%) & \textbf{20.5} & (2.8) & \textbf{76.36\%} & (1.99\%) \\
\bottomrule
\end{tabular}%

\label{tab:ArcadeWeightsEfficientNetB3}
\end{table*}

Finally, with EfficientNetV2S (Table~\ref{tab:ArcadeWeightsEfficientNetV2S}), using the \emph{Arcade} weights led to improved accuracy in only 9 out of the 22 systems. However the average accuracy still increased from $77.44\%$ to $77.63\%$. Training times decreased for 17 of the 22 systems, and the average number of epochs required to train the network decreased from $26.9$ to $24.5$.

\begin{table*}

\caption{Accuracy and epochs required to train the EfficientNetV2S architecture across 22 screenshot datasets, using random weight initialization, ImageNet weights, and Arcade dataset weights as initial weights. Each model is executed five times using the Cross Validation approach. The highest accuracy and the lowest number of epochs for each dataset are highlighted in bold. Standard deviations are shown in parentheses.}

\centering

\begin{tabular}{r|cccc|cccc|cccc}
\toprule
\textbf{Weights} & \multicolumn{4}{c|}{\textbf{Random}} & \multicolumn{4}{c|}{\textbf{ImageNet}} & \multicolumn{4}{c}{\textbf{Arcade}} \\
\textbf{System} & \multicolumn{2}{c}{\textbf{Epochs}} & \multicolumn{2}{c|}{\textbf{Accuracy}} & \multicolumn{2}{c}{\textbf{Epochs}} & \multicolumn{2}{c|}{\textbf{Accuracy}} & \multicolumn{2}{c}{\textbf{Epochs}} & \multicolumn{2}{c}{\textbf{Accuracy}} \\
Atari 2600 & 42.0  & (6.2) & 82.31\% & (1.77\%) & 31.8  & \multicolumn{1}{c}{(3.7)} & \textbf{88.92\%} & (2.74\%) & \textbf{28.8} & \multicolumn{1}{c}{(5.8)} & 88.78\% & (0.82\%) \\
NES   & 23.0  & (0.6) & 60.43\% & (2.11\%) & 20.6  & \multicolumn{1}{c}{(1.9)} & \textbf{72.51\%} & (2.40\%) & \textbf{17.4} & \multicolumn{1}{c}{(0.5)} & 69.33\% & (2.64\%) \\
Master System & 27.8  & (3.0) & 61.61\% & (1.32\%) & 26.0  & \multicolumn{1}{c}{(1.4)} & \textbf{76.20\%} & (1.19\%) & \textbf{24.2} & \multicolumn{1}{c}{(1.2)} & 74.45\% & (0.81\%) \\
PC Engine & 25.6  & (1.5) & 68.58\% & (2.00\%) & 25.6  & \multicolumn{1}{c}{(1.6)} & \textbf{80.52\%} & (0.94\%) & \textbf{23.6} & \multicolumn{1}{c}{(2.6)} & 79.33\% & (0.46\%) \\
Mega Drive & 23.6  & (1.4) & 59.26\% & (1.90\%) & 27.2  & \multicolumn{1}{c}{(1.3)} & \textbf{74.56\%} & (3.22\%) & \textbf{20.0} & \multicolumn{1}{c}{(2.1)} & 73.11\% & (1.87\%) \\
Super Nintendo & 22.0  & (0.6) & 59.10\% & (1.38\%) & 21.6  & \multicolumn{1}{c}{(2.7)} & \textbf{72.89\%} & (1.72\%) & \textbf{17.6} & \multicolumn{1}{c}{(2.7)} & 71.88\% & (2.25\%) \\
Sega Saturn & 27.0  & (3.0) & 56.83\% & (1.96\%) & \textbf{31.0} & \multicolumn{1}{c}{(3.6)} & 75.91\% & (1.11\%) & 31.8  & \multicolumn{1}{c}{(8.2)} & \textbf{83.14\%} & (1.26\%) \\
PlayStation & 23.2  & (1.6) & 59.38\% & (1.88\%) & 22.0  & \multicolumn{1}{c}{(3.0)} & \textbf{74.77\%} & (2.49\%) & \textbf{19.4} & \multicolumn{1}{c}{(2.2)} & 73.62\% & (2.20\%) \\
Nintendo 64 & 31.0  & (1.5) & 67.11\% & (2.21\%) & 33.2  & \multicolumn{1}{c}{(5.3)} & 82.28\% & (3.19\%) & \textbf{26.4} & \multicolumn{1}{c}{(2.9)} & \textbf{85.23\%} & (1.89\%) \\
Dreamcast & 32.6  & (6.8) & 59.80\% & (1.18\%) & 41.2  & \multicolumn{1}{c}{(6.4)} & 74.99\% & (1.83\%) & \textbf{25.6} & \multicolumn{1}{c}{(3.7)} & \textbf{78.86\%} & (1.45\%) \\
PlayStation 2 & 24.2  & (1.2) & 55.53\% & (1.81\%) & \textbf{25.4} & \multicolumn{1}{c}{(1.4)} & 75.14\% & (3.29\%) & \textbf{25.4} & \multicolumn{1}{c}{(2.7)} & \textbf{76.73\%} & (0.63\%) \\
GameCube & 30.0  & (2.0) & 63.74\% & (2.13\%) & \textbf{28.0} & \multicolumn{1}{c}{(2.4)} & 81.48\% & (2.30\%) & 32.2  & \multicolumn{1}{c}{(7.6)} & \textbf{85.00\%} & (0.75\%) \\
Xbox  & 32.2  & (4.7) & 69.38\% & (1.94\%) & 35.0  & \multicolumn{1}{c}{(5.8)} & 84.45\% & (1.32\%) & \textbf{25.8} & \multicolumn{1}{c}{(4.4)} & \textbf{86.42\%} & (1.45\%) \\
Xbox 360 & 27.4  & (2.6) & 60.78\% & (1.36\%) & 25.8  & \multicolumn{1}{c}{(2.3)} & \textbf{80.25\%} & (1.17\%) & \textbf{24.0} & \multicolumn{1}{c}{(1.4)} & 79.01\% & (0.44\%) \\
PlayStation 3 & 27.0  & (1.8) & 51.48\% & (1.03\%) & \textbf{22.8} & \multicolumn{1}{c}{(2.8)} & 71.98\% & (3.04\%) & 23.4  & \multicolumn{1}{c}{(2.2)} & \textbf{72.47\%} & (1.08\%) \\
Wii   & 29.2  & (2.9) & 69.00\% & (1.75\%) & 29.4  & \multicolumn{1}{c}{(8.5)} & \textbf{83.96\%} & (0.69\%) & \textbf{24.6} & \multicolumn{1}{c}{(3.0)} & 83.41\% & (0.53\%) \\
Wii U & 26.8  & (3.2) & 51.87\% & (1.81\%) & 24.8  & \multicolumn{1}{c}{(3.3)} & \textbf{74.97\%} & (2.43\%) & \textbf{20.4} & \multicolumn{1}{c}{(5.1)} & 74.58\% & (3.80\%) \\
PlayStation 4 & 23.8  & (1.6) & 46.96\% & (2.05\%) & 21.2  & \multicolumn{1}{c}{(2.4)} & \textbf{63.17\%} & (2.88\%) & \textbf{20.0} & \multicolumn{1}{c}{(1.8)} & 62.53\% & (1.27\%) \\
Xbox One & 27.2  & (2.7) & 47.73\% & (2.49\%) & 25.0  & \multicolumn{1}{c}{(2.4)} & \textbf{66.25\%} & (2.17\%) & \textbf{22.2} & \multicolumn{1}{c}{(1.9)} & 66.04\% & (1.09\%) \\
Nintendo Switch & 23.0  & (1.7) & 53.22\% & (1.65\%) & 34.2  & \multicolumn{1}{c}{(8.8)} & 80.47\% & (1.70\%) & \textbf{33.2} & \multicolumn{1}{c}{(9.0)} & \textbf{80.59\%} & (1.50\%) \\
Xbox Series X/S & 29.8  & (16.7) & 32.86\% & (21.39\%) & \textbf{15.6} & \multicolumn{1}{c}{(1.7)} & \textbf{100.00\%} & (0.00\%) & 27.8  & \multicolumn{1}{c}{(6.5)} & 94.64\% & (6.59\%) \\
PlayStation 5 & 22.2  & (1.7) & 45.74\% & (2.83\%) & \textbf{23.4} & \multicolumn{1}{c}{(2.9)} & 68.10\% & (2.52\%) & 25.4  & \multicolumn{1}{c}{(5.2)} & \textbf{68.62\%} & (1.74\%) \\
\midrule
\textbf{Average} & 27.3  & (3.1) & 58.30\% & (2.73\%) & 26.9  & (3.4) & 77.44\% & (2.02\%) & \textbf{24.5} & (3.8) & \textbf{77.63\%} & (1.66\%) \\
\bottomrule
\end{tabular}%

\label{tab:ArcadeWeightsEfficientNetV2S}
\end{table*}

Table~\ref{tab:BestOverall} displays the highest accuracy achieved for each system, showcasing the best combination of architecture and initial weights. EfficientNetV2S notably outperforms other architectures, yielding the best results in 15 out of 22 systems, with one tie alongside DenseNet201. EfficientNetB3 excels in six systems, while DenseNet201 is the best in two system, one of them alongside EfficientNetV2S. Concerning the initial weights, the ``Arcade weights'' account for the best results in 12 out of 22 systems, while the other ten systems attained their highest accuracy with the ImageNet initial weights. Contrary to expectations, systems benefiting from the use of arcade weights span multiple generations. This indicates that the complexity of arcade graphics is adequate for producing better initial weights for various generations of home systems.

\begin{table}

\caption{The highest accuracy for each system, achieved with the best combination of architecture and initial weights.}

\centering

\begin{tabular}{rccc}
\toprule
\multicolumn{1}{c}{\textbf{System}} & \textbf{Architecture} & \textbf{Weights} & \textbf{Accuracy} \\
\midrule
Atari 2600 & EfficientNetB3 & Arcade & 90.64\% \\
NES   & EfficientNetV2S & ImageNet & 72.51\% \\
Master System & EfficientNetV2S & ImageNet & 76.20\% \\
PC Engine & EfficientNetB3 & Arcade & 81.08\% \\
Mega Drive & EfficientNetV2S & ImageNet & 74.56\% \\
Super Nintendo & DenseNet201 & ImageNet & 74.51\% \\
Sega Saturn & \makecell{DenseNet201 \\ EfficientNetV2S} & Arcade & 83.14\% \\
PlayStation & EfficientNetV2S & ImageNet & 74.77\% \\
Nintendo 64 & EfficientNetV2S & Arcade & 85.28\% \\
Dreamcast & EfficientNetV2S & Arcade & 78.86\% \\
PlayStation 2 & EfficientNetV2S & Arcade & 76.73\% \\
GameCube & EfficientNetB3 & Arcade & 83.96\% \\
Xbox  & EfficientNetV2S & Arcade & 86.42\% \\
Xbox 360 & EfficientNetV2S & ImageNet & 80.25\% \\
PlayStation 3 & EfficientNetV2S & Arcade & 72.47\% \\
Wii   & EfficientNetV2S & ImageNet & 83.96\% \\
Wii U & EfficientNetB3 & Arcade & 76.52\% \\
PlayStation 4 & EfficientNetB3 & ImageNet & 64.77\% \\
Xbox One & EfficientNetV2S & ImageNet & 66.25\% \\
Nintendo Switch & EfficientNetB3 & Arcade & 81.80\% \\
Xbox Series & EfficientNetV2S & ImageNet & 100.00\% \\
PlayStation 5 & EfficientNetV2S & Arcade & 68.62\% \\
\midrule
\textbf{Average} &       &       & \textbf{78.79\%} \\
\bottomrule
\end{tabular}%

\label{tab:BestOverall}
\end{table}

\section{Limitations}
\label{sec:Limitations}

One of the limitations of the current approach is that there is a separate network for each console system, which means the console system must be known prior to title identification. In future works, this can be addressed in different ways. The most straightforward approach would be to treat each game on each system as a single entity. The drawback is that the resulting network would have more than a hundred thousand outputs. Another approach would be to build a network to detect the console system first, and then another network specialized for that system, similar to those in this paper, would identify the title. The drawback in this scenario is that identifying the system could be challenging, especially with newer systems that do not have distinctive color palettes or graphics complexity limitations like older systems. Finally, a third approach would be to consider each game as an entity regardless of the system, considering that some games are available on multiple systems. A network with two parallel output layers could be used to identify the title and the system simultaneously. Given a screenshot, the probabilities of each system and each title could be used together to identify the title and system among the existing combinations, i.e., combinations of titles and systems that do not exist would be ruled out, and the existing pair with the highest combined probabilities would be chosen. However, the size of the output layer would be almost as large as the first approach.

Another limitation of the presented approach is that adding new game titles requires adding an additional output node to the output layer. Once a node is added, the new connection weights must be learned, so the network would need at least a few more epochs of training. On the other hand, the convolutional weights, especially in the first layers, may not change significantly with the addition of new titles. Therefore, their existing weights could be fine-tuned while the new connections are trained.

\section{Conclusions}
\label{sec:Conclusions}

This paper explores the application of ten distinct CNN architectures (VGG16, ResNet50, ResNet152, MobileNet, DenseNet169, DenseNet201, EfficientNetB0, EfficientNetB2, EfficientNetB3, and EfficientNetV2S) and three transformers architectures (ViT-B16, ViT-L32, and SwinT) for identifying video games through screenshots across 22 diverse home console systems, from Atari 2600 (first released in 1977) to PlayStation 5 (first released in 2020). This is a pioneering work as it is the first attempt to identify video game titles by their screenshots. The experiments confirmed the hypothesis that CNN's inherent capacity of automatically extracting relevant features from images is sufficient to identify video game titles from single screenshots in most scenarios, without relying on other features. Using pre-trained weights from the ImageNet dataset, an average accuracy of $77.44\%$ over all the $22$ systems is achieved with the EfficientNetV2S architecture. It was also the best architecture for 16 of the 22 systems. While transformer architectures consistently outperform CNNs in image classification tasks, this was not the case for identifying video games from screenshots. Although transformers could learn in most scenarios, their average accuracy was lower than that of all CNNs, except for VGG16.

When weights pre-trained on another screenshot dataset (\emph{Arcade}) are used as initial weights instead of those from ImageNet, the accuracy improves for both EfficientNetB3 and EfficientNetV2S, while the number of epochs required to converge decreases. The average accuracy across all 22 systems increases from $77.44\%$ to $77.63\%$ with EfficientNetV2S, and the number of epochs required for convergence decreases from $26.9$ to $24.5$. Although the average accuracy is slightly higher, only nine of the 22 systems actually show improvements. In contrast, EfficientNetB3 shows a more significant improvement, with average accuracy increasing from $74.51\%$ to $76.36\%$, and 19 of the 22 systems showing improvements. Additionally, the number of epochs required for convergence decreases from $24.4$ to $19.9$. Interestingly, DenseNet201 shows improvements in 10 of the 22 systems, but the average accuracy remains the same regardless of whether ImageNet or Arcade weights are used. However, the epochs required for convergence still decrease from 24.6 to 22.0.

Overall, considering only the best architecture and weights for each system, an accuracy of $78.79\%$ is achieved. EfficientNetV2S is responsible for the best results in 15 from the 22 systems. The ``Arcade weights'' are responsible for the best results in 12 of the 22 systems, i.e., starting from weights trained on a different dataset, but the same task, are better than starting with the more general ImageNet weights in most scenarios.

The experiments conducted for this paper provided evidence of the efficacy of CNNs in identifying video games from screenshots. Since the largest EfficientNet architecture explored in this paper achieved the best results, future research will explore even larger CNN architectures and CNN ensembles to further enhance accuracy. The challenges of adding new games to trained networks and identifying a game title without knowing its system a priori should also be investigated, as discussed in Section~\ref{sec:Limitations}. Additionally, this study suggests potential applications in other screenshot-based tasks, such as genre classification or similar game searches, leveraging the efficacy of CNNs in video game identification.

\bibliographystyle{IEEEtran}
\bibliography{videogame}

\begin{IEEEbiography}[{\includegraphics[width=1in,height=1.25in,clip,keepaspectratio]{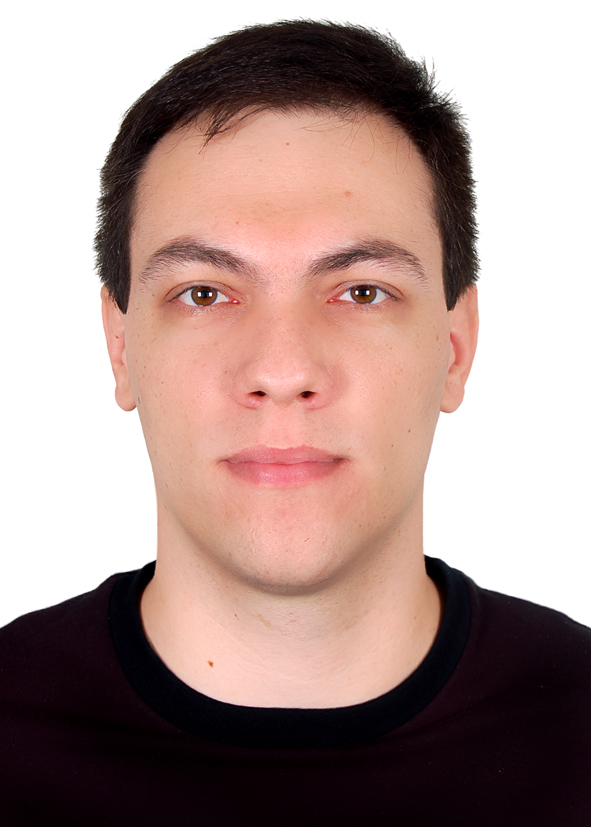}}]{Fabricio Breve}
received his bachelor's degree from the Methodist University of Piracicaba, Brazil in 2001, his master's degree from the Federal University of Sao Carlos, Brazil in 2006, and his Ph.D. from the University of Sao Paulo, Brazil in 2010, with a collaborative period at the University of Alberta, Canada. He is currently an associate professor at Sao Paulo State University, Brazil. His research interests include machine learning, pattern recognition, image processing, artificial neural networks, complex networks, and nature-inspired computing.
\end{IEEEbiography}

\vfill

\end{document}